\documentclass[conference]{IEEEtran}
\IEEEoverridecommandlockouts
\usepackage{cite}
\usepackage{amsmath,amssymb,amsfonts}
\usepackage{algorithmic}
\usepackage{graphicx}
\usepackage{textcomp}
\def\BibTeX{{\rm B\kern-.05em{\sc i\kern-.025em b}\kern-.08em
    T\kern-.1667em\lower.7ex\hbox{E}\kern-.125emX}}

\usepackage[numbers]{natbib}
\usepackage{microtype}
\usepackage{graphicx}

\usepackage{booktabs} 
\usepackage{xcolor}         
\usepackage{colortbl}
\usepackage{graphicx}
\usepackage{subcaption}
\usepackage{comment}
\usepackage[scaled=.8]{beramono}
\usepackage{lipsum}
\usepackage{algorithm}
\usepackage{algorithmic}
\usepackage{multirow}
\usepackage{amsmath}
\usepackage{enumitem}
\usepackage{amssymb}
\usepackage{bm}
\usepackage[subtle]{savetrees}
\usepackage{breakurl}
\usepackage{makecell}
\usepackage{tabularx}
\usepackage{hyperref}
\hypersetup{
  colorlinks=true,
  linkcolor=blue,
  filecolor=magenta,  
  urlcolor=cyan,
  citecolor=blue,
}
\usepackage{tabularx, makecell, booktabs} 

\usepackage[most]{tcolorbox}
\usepackage{algorithm}
\usepackage{algorithmic}  
\usepackage{amsmath}

\definecolor{best}{HTML}{E6F4EA}      
\definecolor{good}{HTML}{FFF4D6}      
\definecolor{poor}{HTML}{FDE0DD}      
\definecolor{thisworkrow}{HTML}{F2FBF2} 

\definecolor{lowcomplex}{HTML}{E6F4EA}
\definecolor{medcomplex}{HTML}{FFF4D6}
\definecolor{highcomplex}{HTML}{FDE0DD}
\definecolor{low}{HTML}{FDE0DD}       
\definecolor{med}{HTML}{FFF4D6}       
\definecolor{medhigh}{HTML}{FFFCEB}   
\definecolor{high}{HTML}{E6F4EA}      
\definecolor{highlightrow}{HTML}{F2FBF2} 

\newcommand{\Fig}[1]{Fig.~\ref{#1}}

\begin{document}


\title{Multi-Agent Reinforcement Learning for \\Sample-Efficient Deep Neural Network Mapping}

\author{
\makebox[\linewidth][c]{%
  \begin{tabular}{ccc}
    Srivatsan Krishnan\textsuperscript{\dag} & Jason Jabbour & Dan Zhang \\
    \textit{NVIDIA} & \textit{Harvard University} & \textit{Google DeepMind}
  \end{tabular}
} \\[3ex]
\makebox[\linewidth][c]{%
  \begin{tabular}{cccc}
    Natasha Jaques\textsuperscript{\dag} & Aleksandra Faust\textsuperscript{\dag} & Shayegan Omidshafiei\textsuperscript{\dag} & Vijay Janapa Reddi \\
    \textit{University of Washington} & \textit{Genesis Therapeutics} & \textit{Field AI} & \textit{Harvard University}
  \end{tabular}
}
}

\maketitle

\begingroup
\renewcommand\thefootnote{\dag}
\footnotetext{Work done while at Google DeepMind.}
\endgroup

\begin{abstract}

Mapping deep neural networks (DNNs) to hardware is critical for optimizing latency, energy consumption, and resource utilization, making it a cornerstone of high-performance accelerator design. Due to the vast and complex mapping space, reinforcement learning (RL) has emerged as a promising approach—but its effectiveness is often limited by sample inefficiency. We present a decentralized multi-agent reinforcement learning (MARL) framework designed to overcome this challenge. By distributing the search across multiple agents, our framework accelerates exploration. To avoid inefficiencies from training multiple agents in parallel, we introduce an agent clustering algorithm that assigns similar mapping parameters to the same agents based on correlation analysis. This enables a decentralized, parallelized learning process that significantly improves sample efficiency. Experimental results show our MARL approach improves sample efficiency by 30–300× over standard single-agent RL, achieving up to 32.61× latency reduction and 16.45× energy-delay product (EDP) reduction under iso-sample conditions.

\end{abstract}



\section{Introduction}

\label{sec:intro}

\begin{table*}[t!]
\centering
\caption{Prior work on DNN mapping. Some approaches achieve high sample efficiency but lack the capability to handle complex, high-dimensional search spaces, while others, like reinforcement learning (RL), are more capable but suffer from sample inefficiency. Our proposed method combines multi-agent RL with clustering to balance these trade-offs, offering both the capability to navigate complex search spaces and the sample efficiency required for practical application.}
\label{tab:prior-work}
\small
\renewcommand{\arraystretch}{1.3}
\resizebox{2\columnwidth}{!}{%
\begin{tabular}{|c|c|c|c|c|c|}
\hline
\textbf{Related Work}  & \textbf{Type}             & \textbf{Algorithm}         & \textbf{Parallelism}  & \textbf{Search Capability} & \textbf{Sample Efficiency \( \bm{\downarrow} \)} \\ \hline\hline
Parashar et al.~\citep{parashar2019timeloop}    & Random Search             & Random Search              & Single Agent         & \cellcolor{low}Low                        & \cellcolor{low}Low \\ \hline
Sunny et al.~\citep{dnn-mapping-rl}             & Machine Learning          & Reinforcement Learning     & Single Agent         & \cellcolor{high}High                       & \cellcolor{low}Low \\ \hline
Shen et al.~\citep{grid-search-dnn1}            & Brute Force               & Grid Search                & Single Agent         & \cellcolor{low}Low                        & \cellcolor{low}Low \\ \hline
Stoutchinin et al.~\citep{grid-serach-dnn2}     & Brute Force               & Grid Search                & Single Agent         & \cellcolor{low}Low                        & \cellcolor{low}Low \\ \hline
Suda et al.~\citep{grid-search-dnn3}            & Brute Force               & Grid Search                & Single Agent         & \cellcolor{low}Low                        & \cellcolor{low}Low \\ \hline
Shi et al.~\citep{bo-dnn-mapping}              & Machine Learning          & Bayesian Optimization (BO) & Single Agent         & \cellcolor{med}Medium                     & \cellcolor{medhigh}High-Medium \\ \hline
Kao et al.~\cite{kao2020gamma}                 & Evolutionary              & Genetic Algorithm (GA)     & Multi-Agent          & \cellcolor{med}Medium                     & \cellcolor{high}High \\ \hline
Kao et al.~\citep{kao2020gamma}                & Evolutionary              & Domain-Specific GA         & Multi-Agent          & \cellcolor{med}Medium                     & \cellcolor{high}High \\ \hline
Hegde et al.~\citep{mindmappings}              & Machine Learning          & Gradient Descent           & Single Agent         & \cellcolor{med}Medium                     & \cellcolor{high}High \\ \hline\hline
\rowcolor{highlightrow}
\textbf{This Work}    & 
\textbf{Machine Learning} & 
\textbf{Clustering + RL} & 
\textbf{Multi-Agent} & 
\cellcolor{high}\textbf{High} & 
\cellcolor{high}\textbf{High} \\ \hline
\end{tabular}}
\vspace{-1em}
\end{table*}


Deep neural networks (DNNs) have become foundational in machine learning, powering transformative models for computer vision~\cite{deeplearning-cv, resnet-cv,mobilenet}, sequential decision-making~\cite{sutton2018reinforcement}, Large Language Models (LLMs)~\cite{llm1, megatron}  and more. Training and deploying these complex DNNs incur substantial computational demands. 
Hardware accelerators like TPUs~\citep{tpu} and GPUs~\citep{choquette2021nvidia} are crucial for feasible deployment, but efficiently mapping DNN layers onto these accelerators remains an open challenge. DNN mapping plays an important role in bridging the gap between computational models and hardware capabilities~\citep{dnn-mapping1, dnn-mapping2}. DNN mapping is the process of optimizing a model's layer configuration to efficiently utilize hardware accelerators (e.g., TPUs, GPUs). By optimally allocating layers to hardware resources, mapping can maximize performance metrics like latency and energy efficiency. 

In this paper, we introduce the first fully-decentralized multi-agent RL framework that is specifically designed to improve the sample efficiency of RL for DNN mapping through parallelized and structured exploration of the design space. RL often relies on trial-and-error exploration, contributing to its sample inefficiency. Gathering each sample requires interactions with the environment, which can be slow and computationally expensive. Given the complexity of DNN model design, with configuration spaces ranging from $\sim$ 10$^{4}$ to 10$^{39}$ per layer, search algorithms need to be sample efficient to navigate this vast design space effectively. This underscores the importance of developing more sample-efficient approaches, such as MARL.


Our key insight is to treat layer mapping decisions as separate agents, where each agent controls one parameter of the action space. While these parameters are interconnected, the agents communicate indirectly through a shared global reward, allowing them to collectively converge on an optimal solution. To address the challenge of fully decentralized agents, where an increase in the number of parameters leads to a proportional rise in the number of agents and potentially increases computational cost, we propose an efficient correlation-based algorithm. Our algorithm assigns highly correlated parameters to the same agent, while distributing less correlated parameters across independent agents. This factorization of the mapping space among decentralized and parallel agents improves the sample efficiency of mapping.

We show that the sample efficiency gains MARL achieves are primarily attributed to efficiently assigning mapping parameters to multiple agents by clustering similar mapping parameters together, and dedicating independent parameters to different agents. This makes the search process orders of magnitude more efficient than single agent RL, where all parameters are assigned to the same agent and that agent must search a combinatorially exploding parameter space.

Our MARL-based mapper demonstrates significantly more sample-efficient convergence compared to single-agent methods, achieving 30-300$\times$ improvement over traditional single-agent reinforcement learning and other baselines.
For instance, when determining the optimal mapping to minimize latency for the second layer of ResNet18, our MARL framework demonstrates significant advantages over traditional single-agent RL. It converges 120× faster in terms of sample efficiency and achieves approximately 1.5× lower latency.
Under an equal sample budget, it also outperforms existing approaches like genetic algorithms (GA) and Bayesian optimization (BO) on objectives like latency, energy-delay product (EDP), and area utilization. 
In general, our results across different DNN models layers and target objectives indicate that our MARL mapper consistently outperforms single-agent RL and other baselines. 

In summary, our contributions are the following:

\begin{itemize}[noitemsep, topsep=0pt, partopsep=0pt, parsep=0pt]

\item We propose a \textbf{fully decentralized MARL framework} for DNN mapping that improves sample efficiency via parallel, structured exploration.

\item We introduce a \textbf{correlation-based agent assignment} algorithm that clusters parameters to effectively factorize the mapping space.

\item Our method achieves \textbf{30–300$\times$ faster convergence} and up to \textbf{1.5$\times$ lower latency} than single-agent RL. 

\item Under equal sample budgets, our approach \textbf{outperforms prior methods} with up to \textbf{6.48$\times$ better latency and EDP}.


\end{itemize}


\section{State-of-the-Art}

Table~\ref{tab:prior-work} shows prior work in DNN mapping, categorized by their underlying search strategies: pruned randomized search, ML, evolutionary algorithms, and brute-force methods.

\textbf{Pruned Search.} Pruned Search. Tools like Timeloop~\citep{parashar2019timeloop} and earlier works~\citep{grid-search-dnn1, grid-serach-dnn2, grid-search-dnn3} rely on randomized or grid-based search to explore layer-specific mappings. However, these methods are sample-inefficient and often miss high-quality mappings, especially as design spaces grow. Our approach overcomes these limitations through more efficient, scalable exploration.


\textbf{Machine Learning.} Techniques such as Bayesian optimization~\citep{bo-dnn-mapping}, gradient descent~\citep{mindmappings}, and reinforcement learning~\citep{dnn-mapping-rl} have also been explored. Yet most rely on single-agent models, which struggle to scale and remain sample-inefficient as search spaces grow. In contrast, our MARL framework factorizes the search space, improving scalability and efficiency.


\textbf{Evolutionary Algorithms.} Genetic algorithms have been applied to DNN mapping, with GAMMA~\citep{kao2020gamma} using domain-specific operators to boost sample efficiency. While evolutionary methods offer parallelism and improved efficiency over random search, they are generally less effective than RL. Our MARL approach demonstrates superior exploration and solution quality.

\section{Seach Space Complexity}
\label{sec:mapping_choices}


We present the sources of complexity in mapping DNN layers to hardware accelerators, their interplay across  architectures, and the resulting combinatorial explosion of mapping choices. 

\subsection{Mapping Choices}
\label{sec:mapping_choices}

\begin{figure*}[t!]
  \centering
\begin{subfigure}[t]{0.65\columnwidth}
  \centering
\includegraphics[width=0.95\columnwidth]{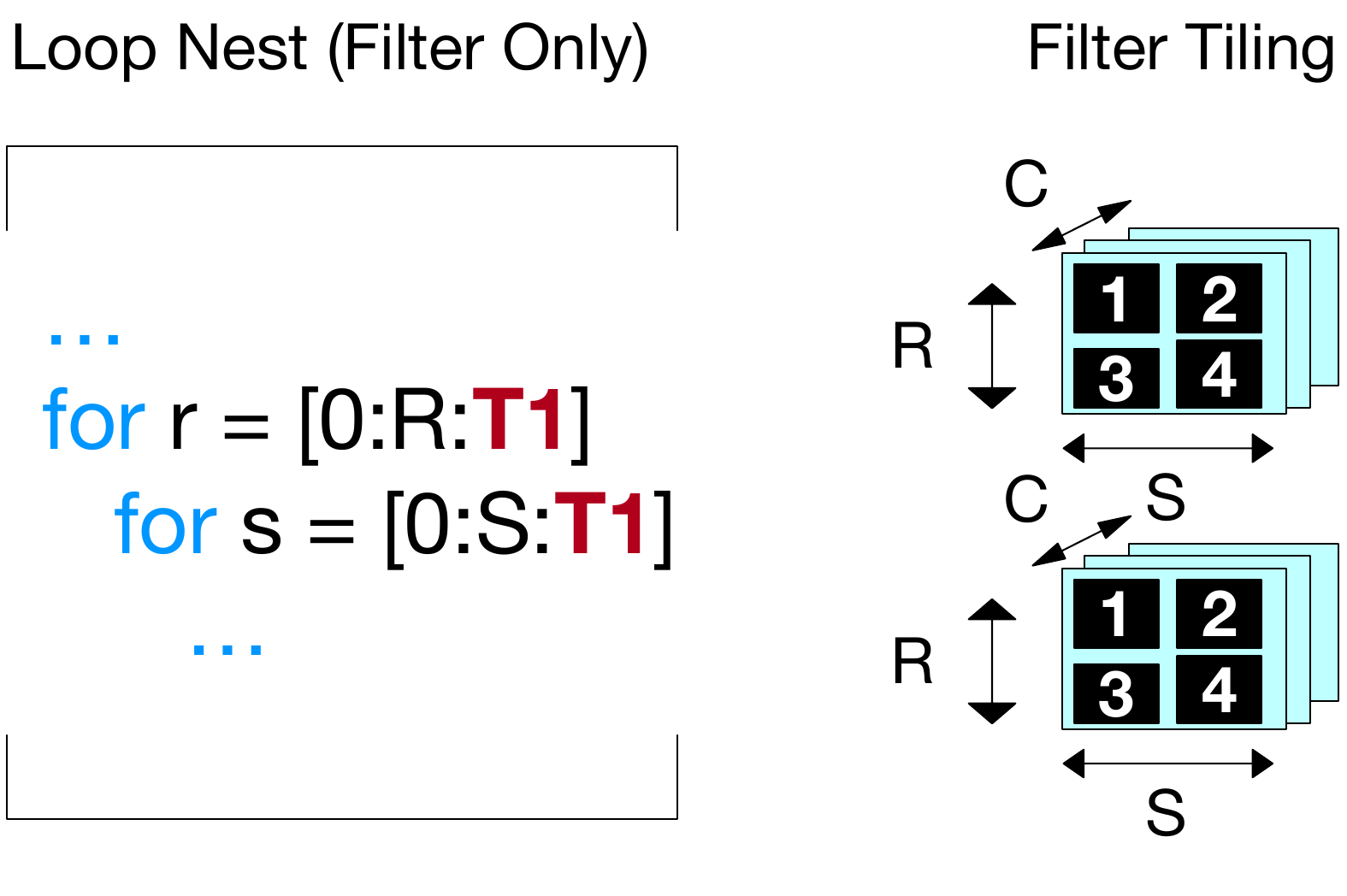}
  \caption{Tiling}
  \label{fig:tiling}
\end{subfigure}%
\begin{subfigure}[t]{0.5\columnwidth}%
  \centering
  \includegraphics[width=0.8\columnwidth]{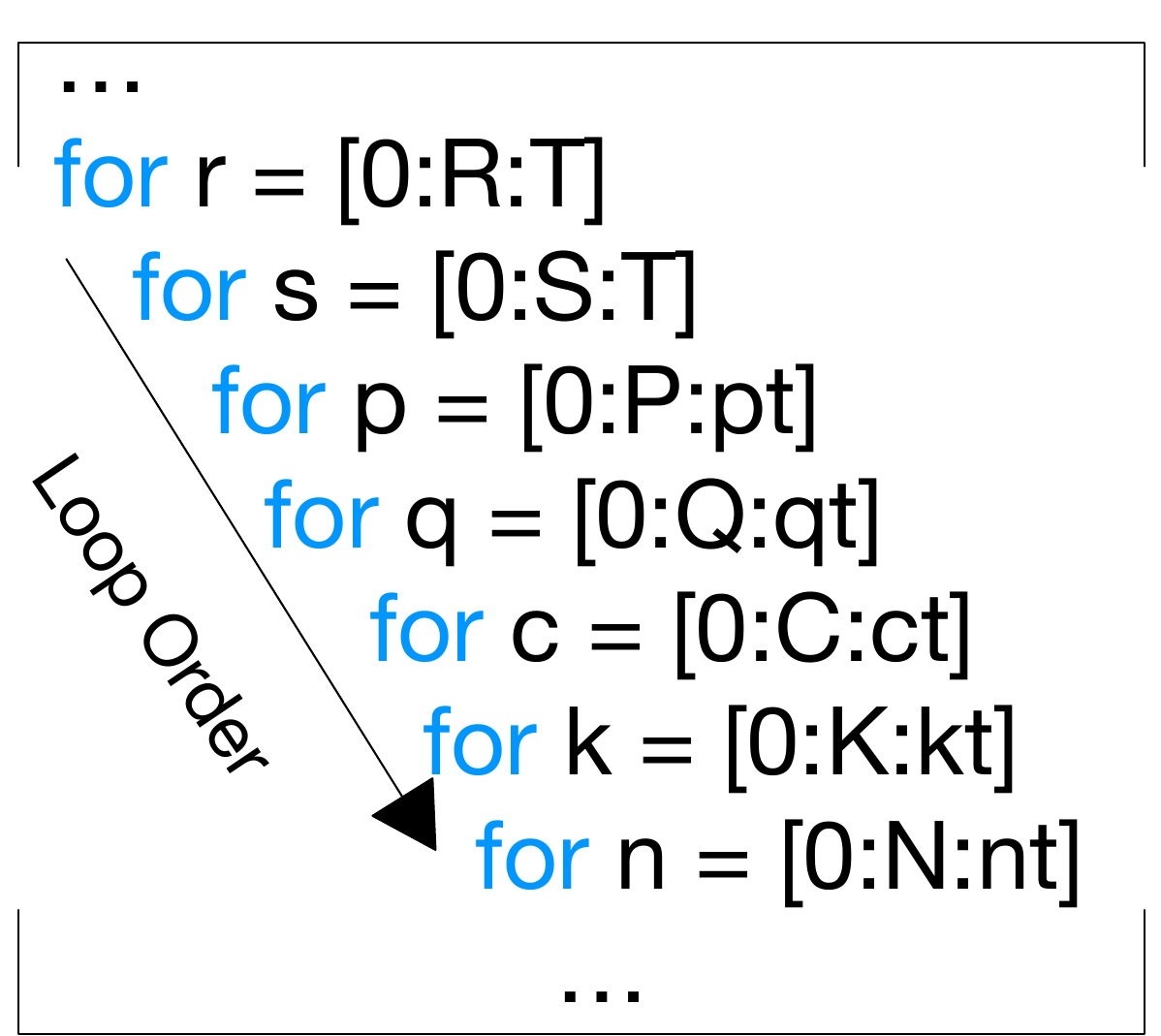}
  \caption{Loop order.}
  \label{fig:loop-order}
\end{subfigure}%
\begin{subfigure}[t]{0.8\columnwidth}%
  \centering
  \includegraphics[width=0.8\columnwidth]{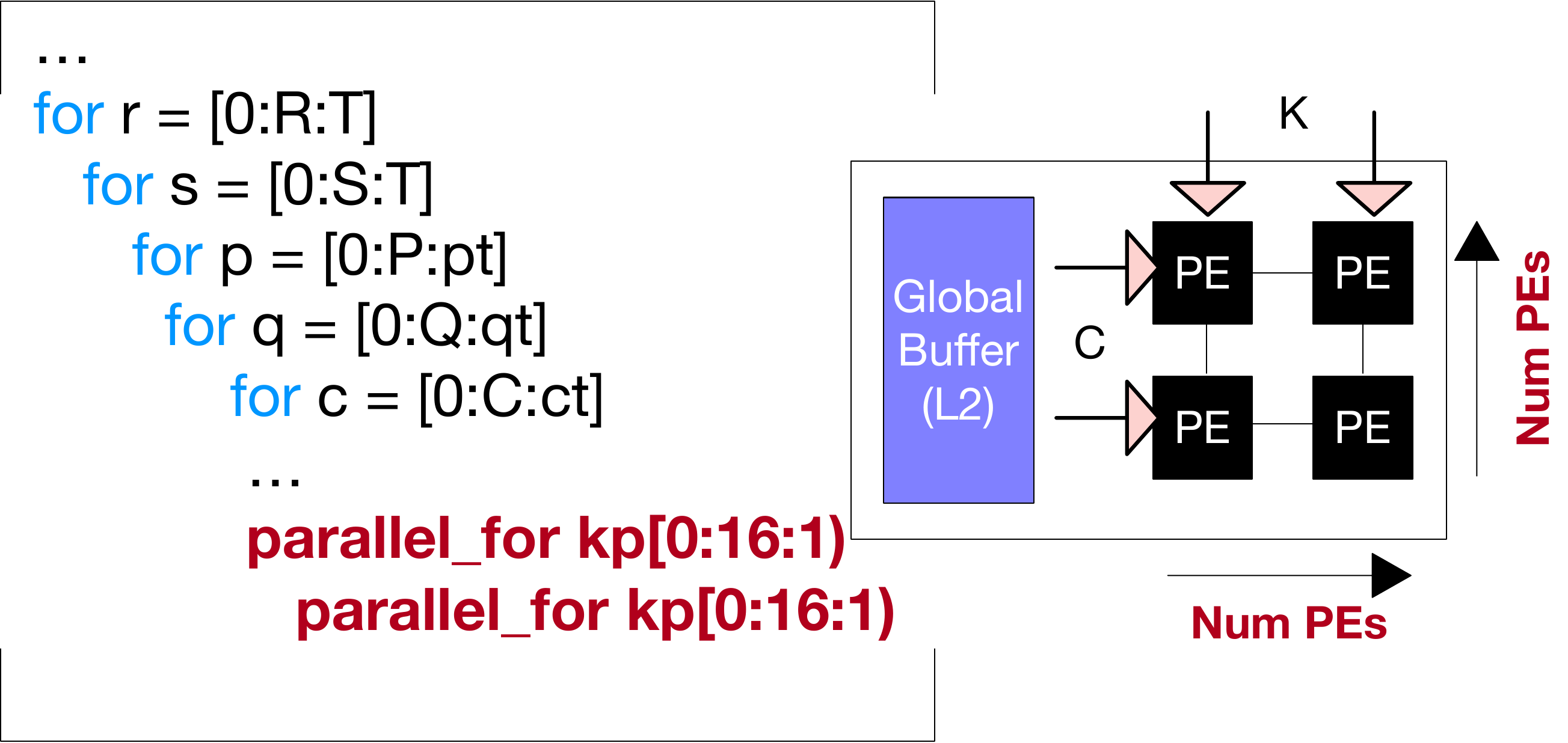}
  \caption{Loop parallelization.}
  \label{fig:loop-parallel}
\end{subfigure}%
\caption{Mapping choices available for a loop nest on a given hardware. (a) Tiling changes the bounds of the loop nest and blocks of various tensors. (b) Loop order iterated  per tiling level. (c) Loop parallelization dimension and number of PEs.  The combination of these mapping choices can result in a large design space (e.g., $\sim$10$^{4}$ to 10$^{39}$ per layer).
\vspace{-1em}
}
\label{fig:mapping}
\end{figure*}

A DNN layer is modeled as a loop nest, with many possible mappings to hardware. As shown in \Fig{fig:mapping}, performance depends on how data is tiled, ordered, and parallelized across memory hierarchies (e.g., DRAM, L2, L1). 

\noindent\textbf{Tiling.} Tiling partitions tensors to fit within hardware memory. As shown in \Fig{fig:tiling}, convolution layers can be tiled across filter dimensions \texttt{\textlangle R, S\textrangle} and channels \texttt{C}. Optimal tiling improves memory, latency, and energy. The search space scales as $n^d$ for $d$ dimensions and grows to $n^{d \times l}$ across $l$ memory levels, yielding billions of configurations.


\noindent\textbf{Loop Order.} Loop order defines the sequence of nested loops in DNN computation, affecting latency by leveraging data locality. As shown in \Fig{fig:loop-order}, a 2D convolution involves seven loops (e.g., \texttt{C}, \texttt{K}, \texttt{\textlangle R, S\textrangle}, \texttt{P}, \texttt{Q}, \texttt{N}), yielding $7!$ orders. With $l$ memory levels, the space grows to $(d!)^l$, reaching ~25M for two levels and ~128B for three—making manual exploration infeasible.

\noindent\textbf{Loop Parallelization.} Loop parallelization distributes computation across processing elements (PEs). As shown in \Fig{fig:loop-parallel}, dimensions like \texttt{K} and \texttt{C} can be split across PEs to boost throughput. The number of configurations grows combinatorially as $\frac{d!}{(d - k)!}$, where $d$ is the number of loops and $k$ the levels of parallelism.


\subsection{Combined Search Space}
\label{sec:motivation}

Analyzing tiling, loop order, and parallelization reveals the complexity of DNN mapping, but their interaction leads to a combinatorial explosion. The overall search space, shown in Equation~\ref{eq:search_space_complexity}, combines tiling ($n^d$), loop order ($d!$), and parallelization ($\frac{d!}{(d-k)!}$), amplified across $l$ memory levels:

\vspace{-0.5em} \begin{equation} \text{Search Space Complexity} = \left( n^d \times d! \times \frac{d!}{(d-k)!} \right)^l \label{eq:search_space_complexity} \end{equation}

This complexity depends on $n$ and $d$, and thus varies with the DNN operation. To compare architectures, we examine their core operations: linear layers in MLPs, convolutions in CNNs, and attention in Transformers. As shown in Table~\ref{tab:search-space}, CNNs exhibit the largest search space among the three. Compared to Transformers, CNNs have more complex loop structures and data reuse patterns, making them significantly harder to optimize from a design space exploration perspective.

Therefore, in this work, we use CNNs as the primary example to demonstrate our MARL-based framework. While CNNs present the largest search space complexity, the proposed approach is broadly applicable to any DNN mapping problem, including those involving MLPs and Transformers.

\begin{table}[b!]
\centering
\caption{Search space complexity of layer types.}
\label{tab:search-space}
\small
\renewcommand{\arraystretch}{1.3}
\setlength{\tabcolsep}{4pt}

\begin{tabularx}{\columnwidth}{|l|>{\raggedright\arraybackslash}p{2cm}|>{\raggedright\arraybackslash}X|c|}
\hline
\thead{Architecture} & \thead{Operation} & \thead{Parameters} & \thead{Search\\Space\\Complexity} \\ \hline\hline

\rowcolor{lowcomplex}
\textbf{MLPs} & Linear Layer & Input ($I$), Output ($O$), Batch ($B$) & $\sim 10^4$–$10^{13}$ \\

\rowcolor{medcomplex}
\textbf{Transformers} & Attention Mechanism & Sequence length ($T$), Embedding ($C$), Head size ($H$), Heads ($N$), Batch ($B$) & $\sim 10^8$–$10^{26}$ \\

\rowcolor{highcomplex}
\textbf{CNNs} & Convolutional Layer & Input size ($H \times W$), Channels ($C$), Filter size ($R \times S$), Output ($K$), Batch ($B$) & $\sim 10^{13}$–$10^{39}$ \\

\hline
\end{tabularx}
\vspace{-1em}
\end{table}

\section{Multi-Agent Framework}

To address the large search space complexity, we propose a Multi-Agent Reinforcement Learning (MARL) approach for efficient DNN mapping. To address the large search space complexity, we propose a Multi-Agent Reinforcement Learning (MARL) approach for efficient DNN mapping.

\subsection{DNN Mapping as Interconnected Components}


DNN models are viewed as groups of layers with input, weight (or filter), and output tensors, each multi-dimensional, as shown in Fig. \ref{fig:interconnected-systems}a. This has been the basis for many DNN mapping strategies, treating each layer as a single entity with a set of parameters to optimize. In contrast, we propose viewing DNN mapping as a network of interconnected components with interlinked parameters (Fig. \ref{fig:interconnected-systems}b). This perspective introduces an interesting agent assignment problem and better captures complex interactions between the DNN architecture and hardware.

\subsection{Agent Assignment in the Interconnected Model}

The interconnected view of DNN mapping enables flexible agent assignment. On one end, a single RL agent manages all parameters; on the other, each parameter is controlled by a dedicated agent, cooperating toward a shared goal. Between these extremes, a smarter strategy clusters related parameters under shared agents, balancing coordination and learning complexity.

\begin{figure}[h!]
    \centering
    \includegraphics[width=1.0\columnwidth]{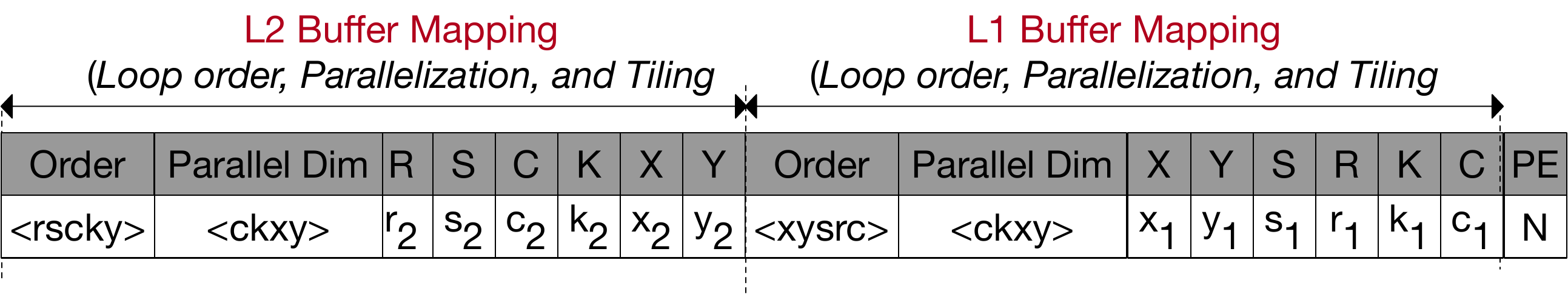}
    \caption{Mapping choices for a given DNN layer assuming there are two levels of memory hierarchy in the accelerator.}
    \label{fig:mapping_options}
\end{figure}

\textit{Single-agent RL.}
A single agent receives the state \texttt{\textlangle latency, power, energy, area\textrangle} and outputs all mapping decisions (\Fig{fig:mapping_options}). For 10 parameters with 10 settings each, this results in a search space of $10^{10}$ combinations. The agent must explore all possible interdependencies, leading to a larger number of exploration steps and reduced sample efficiency.

\textit{Fully Decentralized Learning.}
In this multi-agent setup, each parameter is controlled by a separate RL agent with shared input and individual output. Assuming parameter independence, the 10-parameter space can be explored in 10 steps instead of $10^{10}$. However, this requires separate training resources for each RL agent but significantly reduces the overall exploration needed.

\begin{figure*}[t!]
    \centering
    \hspace{30pt}
    \includegraphics[width=1.9\columnwidth]{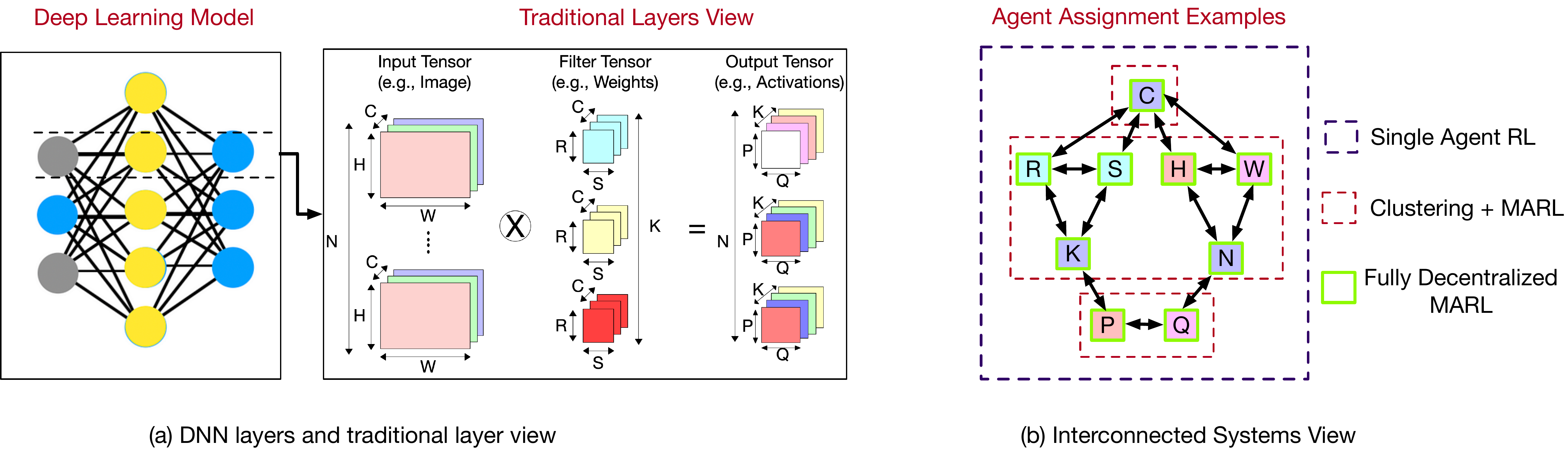}
    \caption{(a) Traditional view: DNN models are represented as layers with input, weight (filter), and output tensors, each treated as independent entities with their own optimization parameters. (b) Interconnected systems view: We propose treating DNN mapping as a network of interlinked components. Examples include single-agent (all parameters assigned to one agent), clustering + MARL (related parameters grouped by agent), and fully decentralized (each parameter assigned to a unique agent).
    \vspace{-1em}}
    
    \label{fig:interconnected-systems}
\end{figure*}


\textit{Clustered Decentralized Learning.}
This clustered approach represents a balanced multi-agent setup. By grouping correlated parameters under shared agents, we address the sample inefficiency of single-agent RL and the training overhead of fully decentralized MARL. This strategy reduces the number of agents while preserving key parameter interactions. For example, in \Fig{fig:interconnected-systems}b, tiling parameters \texttt{R, S, K, H, W, N} can be assigned to one agent, while \texttt{P, Q}, and \texttt{C} are managed by others. This setup improves sample efficiency without incurring the full cost of decentralization.



\subsection{MARL-Based DNN Mapper with Clustering} 
\label{sec:marl_method}

Building on this perspective of interconnected systems, we propose two key innovations: (1) a \textbf{MARL-based DNN Mapper}, which, to our knowledge, represents the first application of Multi-Agent Reinforcement Learning to the DNN mapping problem. This novel approach leverages the flexibility of the interconnected view, enabling a more adaptive and efficient mapping strategy. (2) A \textbf{Clustering Algorithm for Agent Assignment}, where we introduce a new method to efficiently assign agents in a decentralized MARL system for DNN mapping. This algorithm analyzes relationships between mapping parameters to identify optimal groupings for agent control.


\subsection{Algorithm Details} 

\begin{tcolorbox}[colback=gray!10!white, colframe=gray!80!black, boxrule=0.1pt, arc=0pt]
\begin{algorithm}[H]
\caption{Agglomerative clustering for agent asgmt.}\label{alg:agent_assignment}
\begin{algorithmic}[1]
\small
    \REQUIRE Layer dimensions $L = [r,s,k,c,h,w]$, where each $L$ element is an integer, and agent budget $B$.
    \ENSURE Agent assignment index $I$
    
    \STATE \textbf{Step 1:} Initialize layer dimensions $L = [r,s,k,c,h,w]$.
    
    \STATE \textbf{Step 2:} Collect a dataset pairs \texttt<parameters, reward\texttt> $D$ using policy $\pi$.
    
    \STATE \textbf{Step 3:} Construct the correlation matrix $M$ for dataset $D$:
    \[
        M_{ij} = \text{Corr}(D_i, D_j) \quad \text{for } i,j = 1,\ldots,N,
    \]
    where $N$ is the number of parameters, and $\text{Corr}(D_i, D_j)$ calculates the similarity between parameters $D_i$ and $D_j$.
    
    \STATE \textbf{Step 4:} Perform agglomerative clustering on $M$ to obtain $B$ clusters $C_1, C_2, \ldots, C_B$.
    
    \STATE \textbf{Step 5:} Assign indices to dedicated agents based on clustering results:
    \[
        I_i = j \quad \text{if } L_i \in C_j, \quad \text{for } i=1,\ldots,N,
    \]
    where $I_i$ is the assigned agent index for parameter $L_i$ and $j$ is the cluster index.
\end{algorithmic}
\end{algorithm}
\end{tcolorbox}

To systematically assign DNN mapping parameters to agents, we introduce a clustering-based algorithm. It operates in two phases: (1) cluster related parameters based on their correlation with each other and the target objective, and (2) assign each cluster to a dedicated agent. This enables decentralized MARL to control each group independently.

\textbf{Input} The algorithm optimizes parameter control across \texttt{B} agents, based on:

\textit{Layer Configuration:} A vector \emph{L = [r, s, k, c, h, w]} representing DNN layer parameters.

\textit{Agent Budget:} An integer \texttt{B}, the number of available agents.





\textbf{Process.} The algorithm proceeds through these steps:

\textit{Data Acquisition:} Acquire a dataset \texttt{D} through a designated policy $\pi$. The policy $\pi$ can take various forms to explore the parameter space effectively. It may employ heuristics or pruned random search techniques, as described in~\citep{parashar2019timeloop} and~\citep{grid-search-dnn1}. Alternatively, machine learning algorithms such as Bayesian optimization can be utilized. Evolutionary algorithms, like those presented in~\citep{kao2020gamma}, offer another approach to policy implementation. Lastly, single-agent reinforcement learning can also serve as a viable policy option. Each of these methods provides unique advantages in navigating the complex landscape of DNN mapping parameters.     The dataset consists of \texttt{\textlangle parameters, reward\textrangle} pairs collected during various exploration phases with policy $\pi$.

\begin{figure*}[ht!]
\centering
\hspace{-1em}
\begin{subfigure}[t]{1.0\columnwidth}%
  \centering
  \includegraphics[width=\columnwidth]{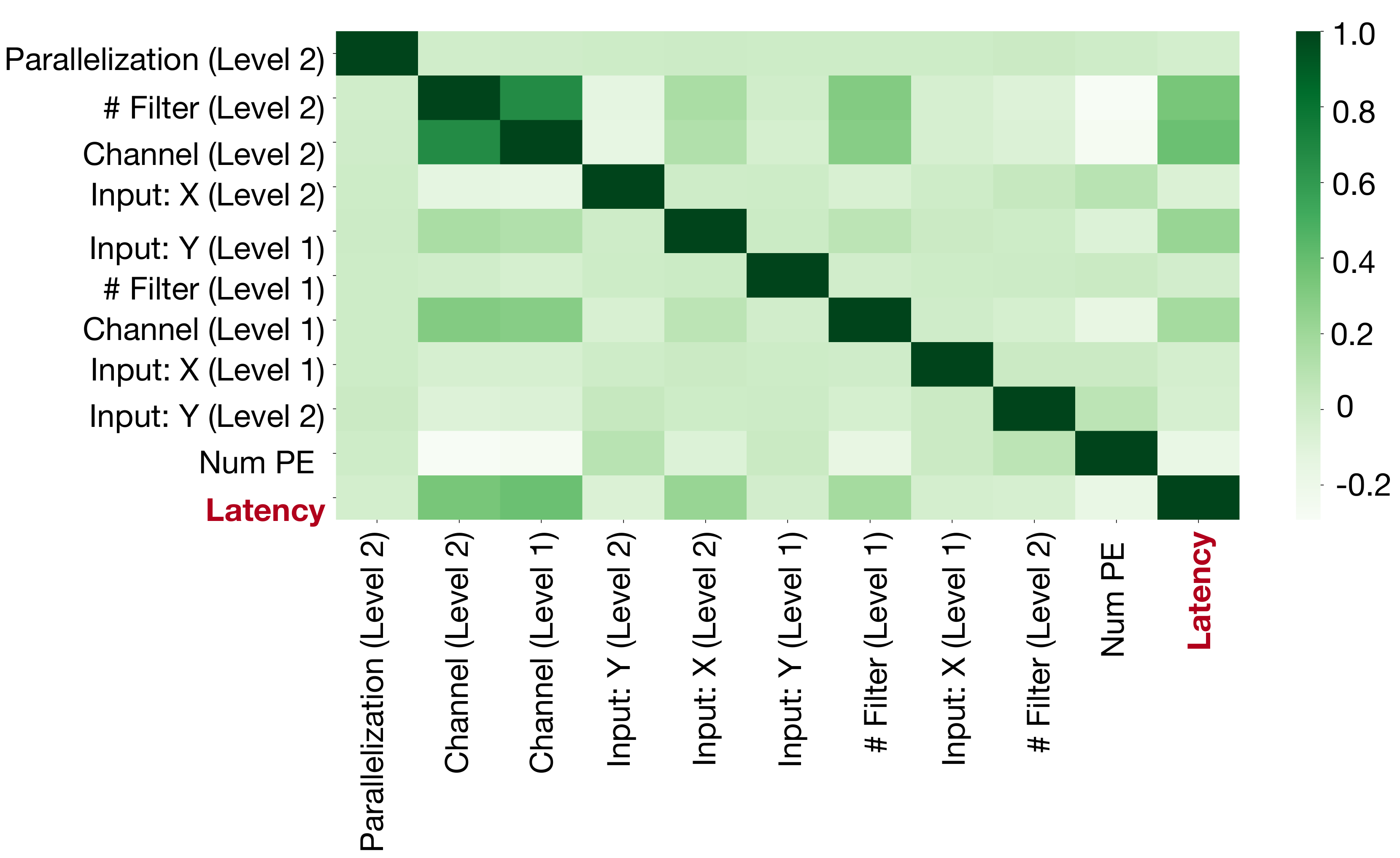}
  \caption{MobileNet Latency.}
  \label{fig:corr-mobilenet}
\end{subfigure}%
\hfill 
\begin{subfigure}[t]{0.7\columnwidth}
  \centering
  \includegraphics[width=\columnwidth]{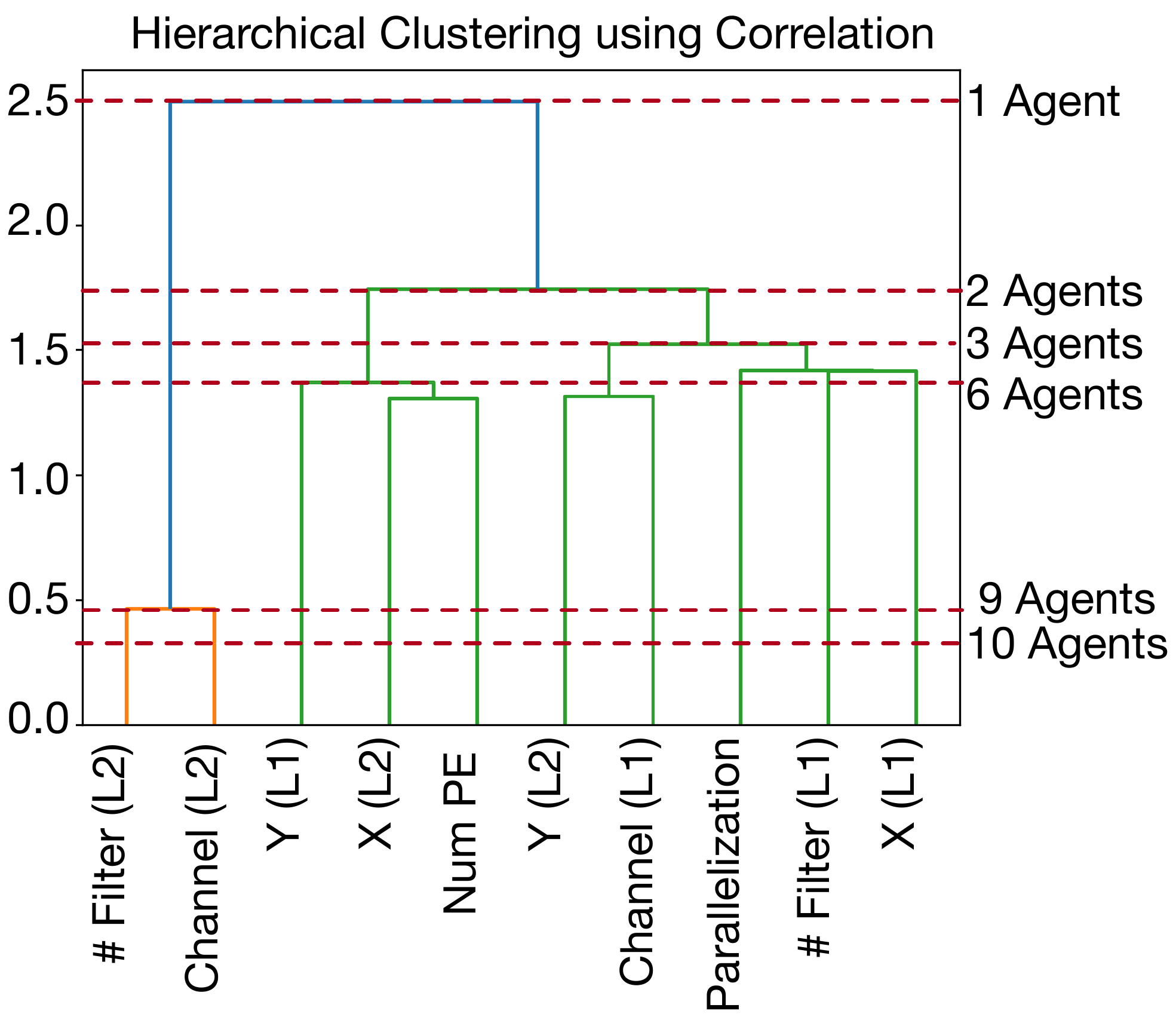}
  \caption{Dendogram for MobileNet Latency.}
  \label{fig:dendo-mobilenet}
\end{subfigure}%
\caption{MARL-based clustering algorithm uses a correlation matrix to identify related parameters for a given objective (e.g., Latency). (a) Correlation matrix for MobileNet-v2 layer 2, constructed from exploration data collected using policy $\pi$ (e.g., random walker, genetic algorithm). (b) We perform agglomerative clustering~\citep{agglo-clustering} to find the similarity between various DNN parameters.} 
\label{fig:correlation}
\end{figure*}

\textit{Data Selection:} Select the top $n\%$ of reward-achieving configurations (i.e., those with the lowest latency, energy-delay product, and area utilization) to counter the sub-optimal exploration when using a policy $\pi$. In our implementation, we set $n = 15$, balancing between exploration breadth and computational efficiency.

\textit{Correlation Analysis:} Compute the correlation matrix (\texttt{M}) using Pearson's $r$ to uncover relationships among the parameters.

\textit{Clustering:} Use the agglomerative clustering algorithm~\citep{agglo-clustering-2} to construct a dendrogram based on pairwise correlations between parameters.

\textit{Agent Assignment:} Choose a threshold based on the agent budget \texttt{B} to group parameters:
    \begin{itemize}
        \item If \texttt{B} = 10 (fully decentralized MARL): Choose threshold $<$0.5, assigning each parameter to a single agent.
        \item If \texttt{B} = 1 (single-agent RL): Choose threshold $>$ 2.5, assigning all parameters to a single agent.
    \end{itemize}

\textit{Role Assignment:} Assign each parameter $L_i$ a dedicated agent role based on its cluster membership.

\textbf{Example.} We present an example of the correlation matrix for the MobileNet-v2 layer 2 model in \Fig{fig:corr-mobilenet}. 
The pairwise correlation between different parameters is utilized by the agglomerative clustering algorithm~\citep{agglo-clustering-2} to construct the dendrogram as depicted in \Fig{fig:dendo-mobilenet}. Depending on the agent budget \texttt{B}, the threshold is selected to group certain parameters (based on their similarity) and assign them to each agent. For example, with an agent budget of 10, a threshold of $<$0.5 is chosen, whereby each parameter is assigned to a single agent (i.e., fully decentralized MARL). In contrast, if the agent budget \texttt{B} is 1, a threshold of $>$ 2.5 is selected, resulting in the assignment of all parameters to a single agent (i.e., single-agent RL). Each parameter $L_i$ is designated a specific agent role based on its membership in the cluster, offering a methodical agent assignment approach.

\section{Experimental Setup}
\label{sec:eval}

We begin by outlining the workload and target objectives, detailing the specific DNN models and layers used as inputs, and explaining our focus on minimizing latency, Energy Delay Product (EDP), and area utilization. Next, we describe the construction of our DNN Mapping Gym Environment, built upon the Maestro infrastructure, which provides a flexible and architecture-agnostic platform for our experiments. 


\subsection{Workloads and Target Objectives} 

In our evaluation, we focus on CNNs since they present the most complex search space (as shown previously in Table~\ref{tab:search-space}), making them the ideal candidate to demonstrate the strength of our approach. However, it is important to note that the techniques we develop are not specific to CNNs. The principles presented here can be applied to other neural network architectures, such as MLPs and Transformers.

The input to the mapper consists of DNN models composed of layers. A layer is passed to the decentralized multiagent mapper, along with the baselines. For each layer in a DNN model topology, the mapper's goal is to minimize the target objectives, namely latency (which affects performance), Energy Delay Product (EDP) that influences both performance and energy, and area utilization. We sample the layers from prominent CNN models, including MobileNet-v2~\citep{mobilenet-v2}, ResNet18~\citep{resnet-cv}, VGG16~\citep{vgg}, and AlexNet~\citep{alexnet}. These layers are selected for their high complexity in the design space, showcasing the effectiveness of our mappers in navigating extensive design spaces to discover the optimal mapping configuration.

\subsection{Gym Environment}
\paragraph{Environment} We build on Maestro~\citep{dnn-mapping1}, a fast, open-source framework for modeling DNN layer mappings across diverse accelerator types (e.g., ASICs, FPGAs). Its hardware-agnostic design enables broad applicability. We extend Maestro into a Gym-compatible RL environment~\cite{gym,archgym}, named \texttt{MaestroEnv}. This encapsulates the DNN mapping task in a standard RL interface, allowing agents to explore dataflow strategies and receive feedback in a structured manner.

\begin{figure*}[t!]
    \centering
    \begin{subfigure}[t]{0.305\textwidth}
        \centering
        \includegraphics[width=\textwidth]{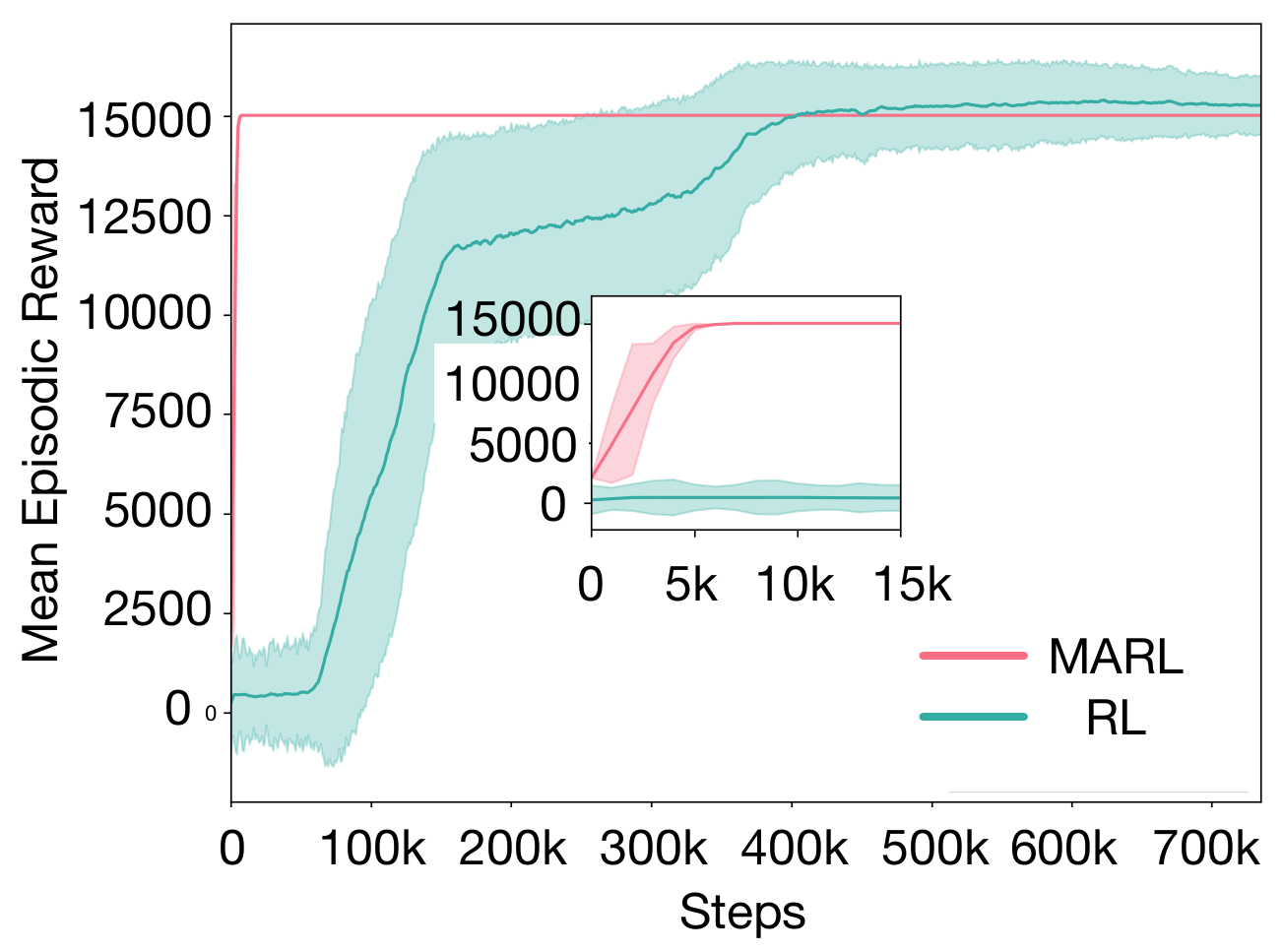}
        \caption{MobileNet-v2 Layer 2 Latency.}
        \label{fig:mobilenet-latency}
    \end{subfigure}
    \hspace{10pt} 
    \begin{subfigure}[t]{0.305\textwidth}
        \centering
        \includegraphics[width=\textwidth]{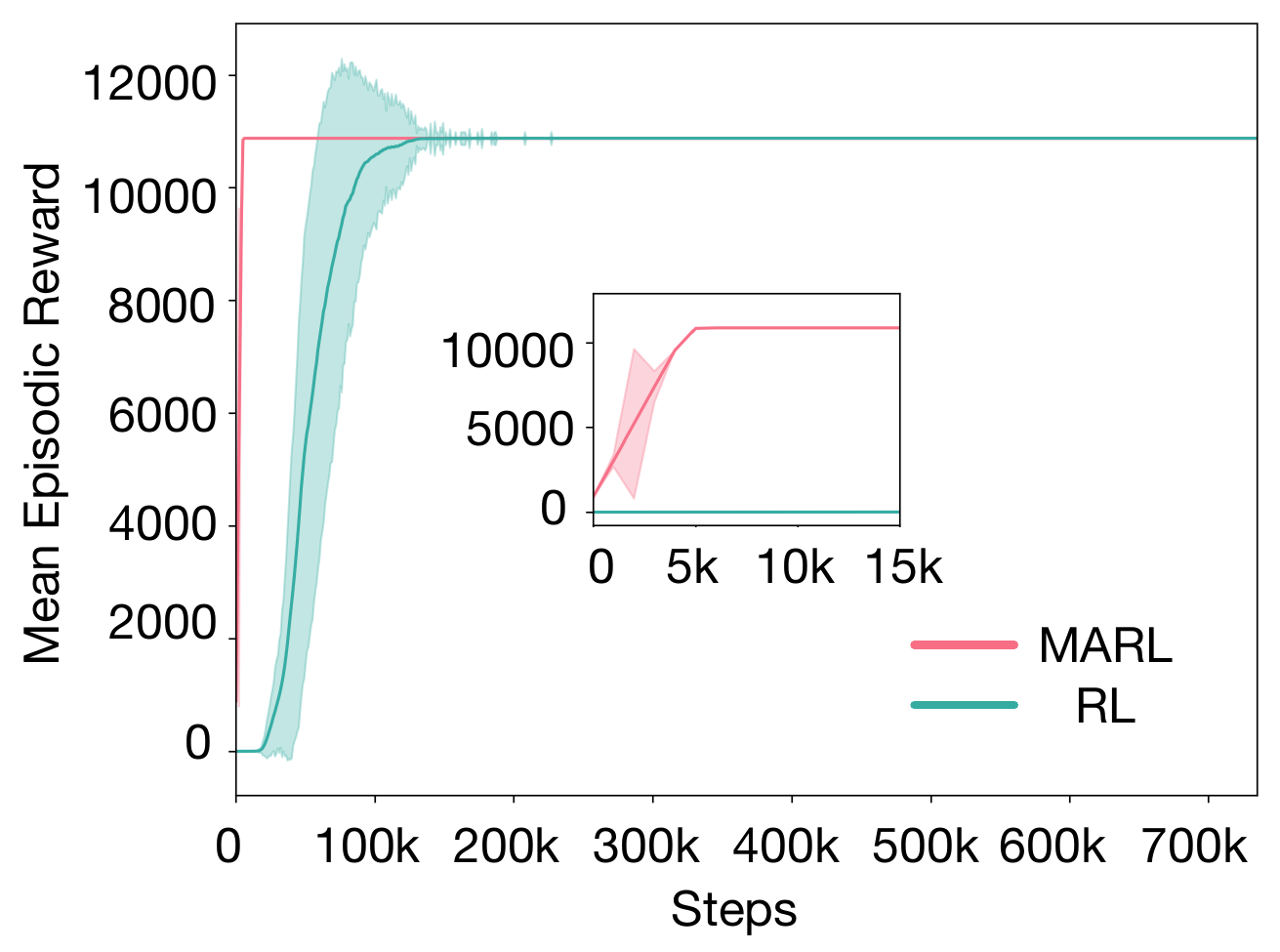}
        \caption{MobileNet-v2 Layer 2 Area.}
        \label{fig:mobilenet-area}
    \end{subfigure}
    \hspace{10pt}
    \begin{subfigure}[t]{0.32\textwidth}
        \centering
        \includegraphics[width=\textwidth]{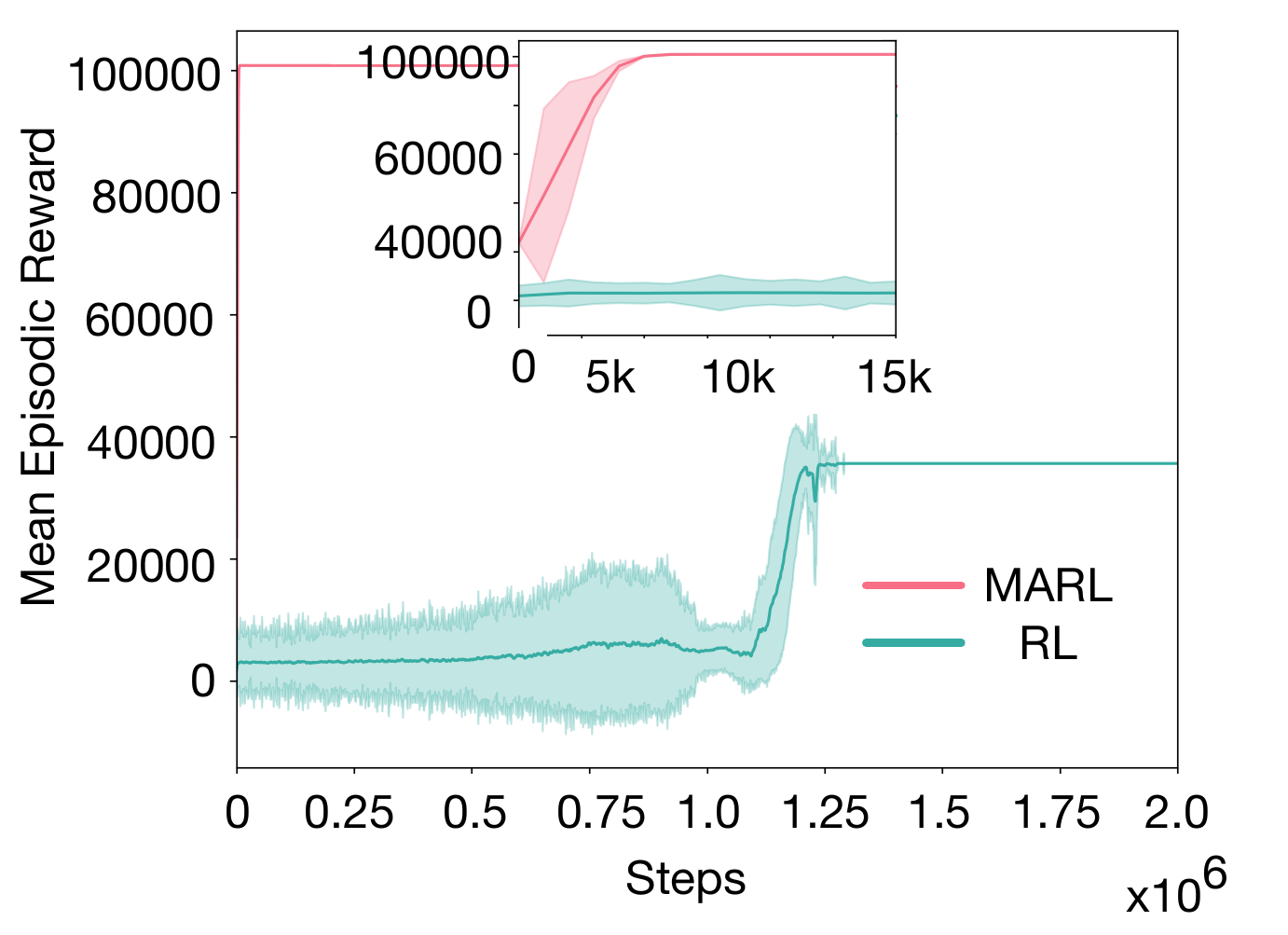}
        \caption{MobileNet-v2 Layer 2 EDP.}
        \label{fig:mobilenet-energy}
    \end{subfigure}

    \vspace{5pt} 
    \begin{subfigure}[t]{0.3\textwidth}
        \centering
        \includegraphics[width=\textwidth]{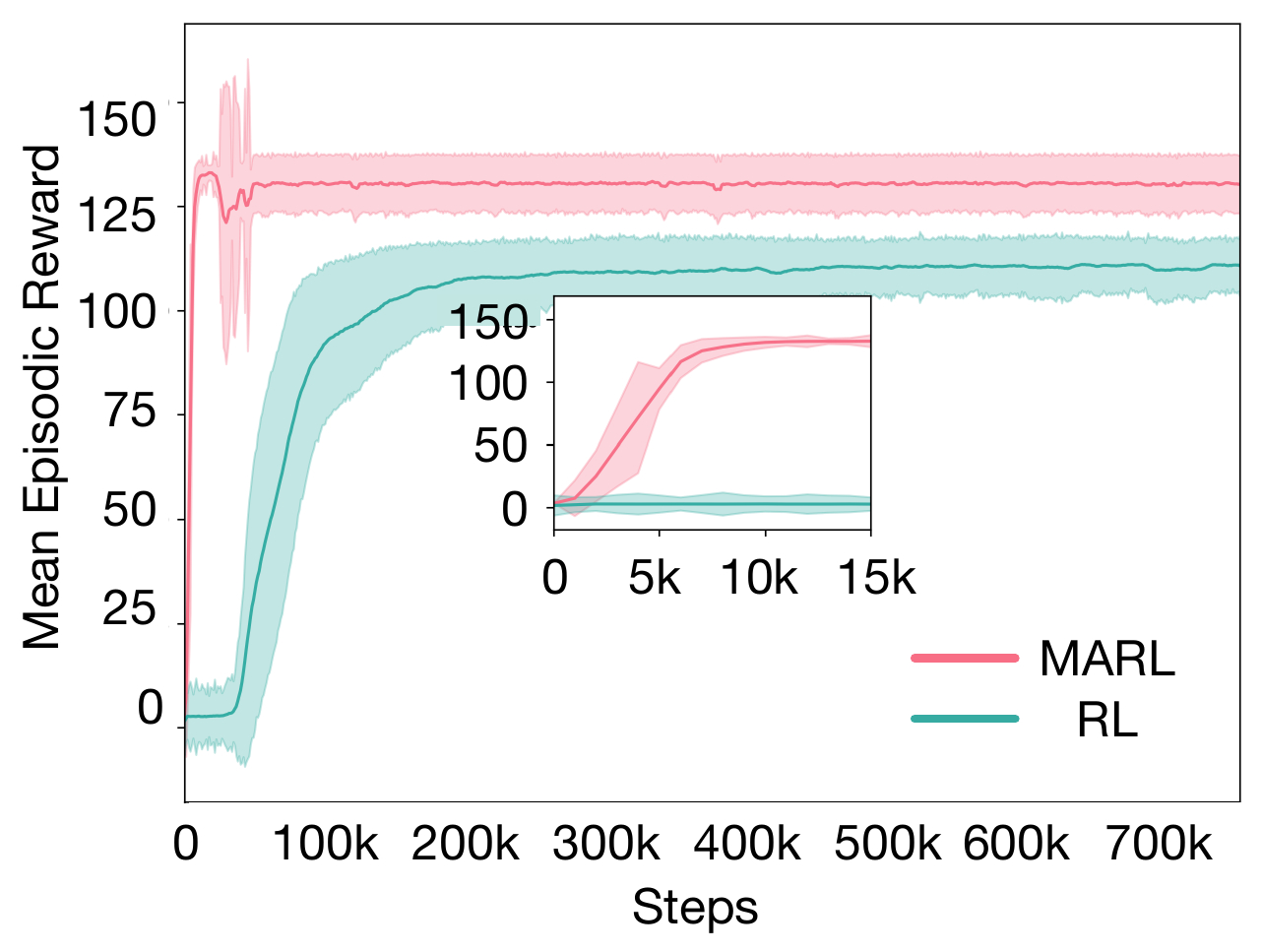}
        \caption{VGG16 Layer 2 Latency.}
        \label{fig:vgg-latency}
    \end{subfigure}
    \hspace{10pt}   
    \begin{subfigure}[t]{0.31\textwidth}
        \centering
        \includegraphics[width=\textwidth]{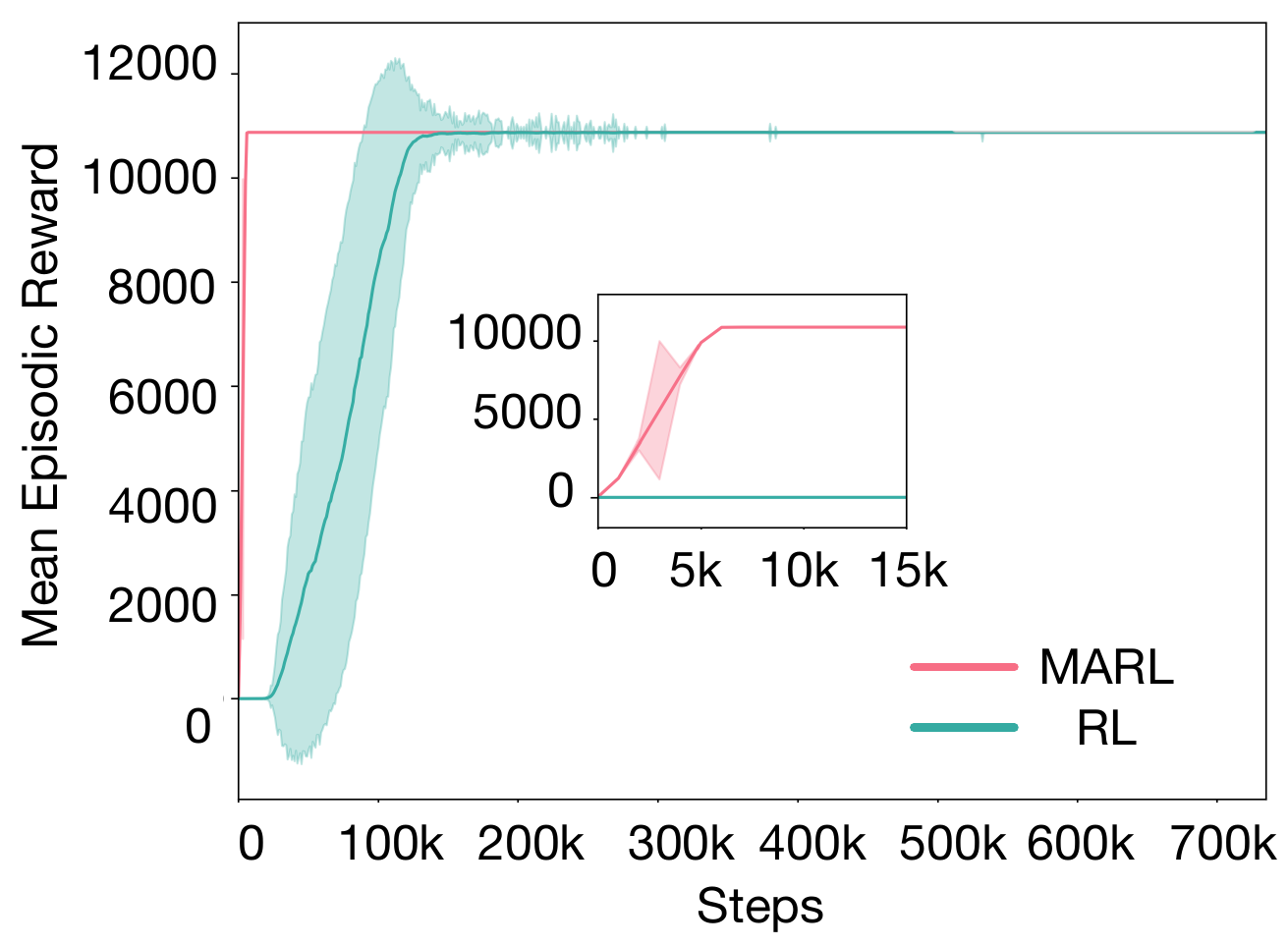}
        \caption{VGG16 Layer 2 Area.}
        \label{fig:vgg-area}
    \end{subfigure}
    \hspace{10pt}
    \begin{subfigure}[t]{0.3\textwidth}
        \centering
        \includegraphics[width=\textwidth]{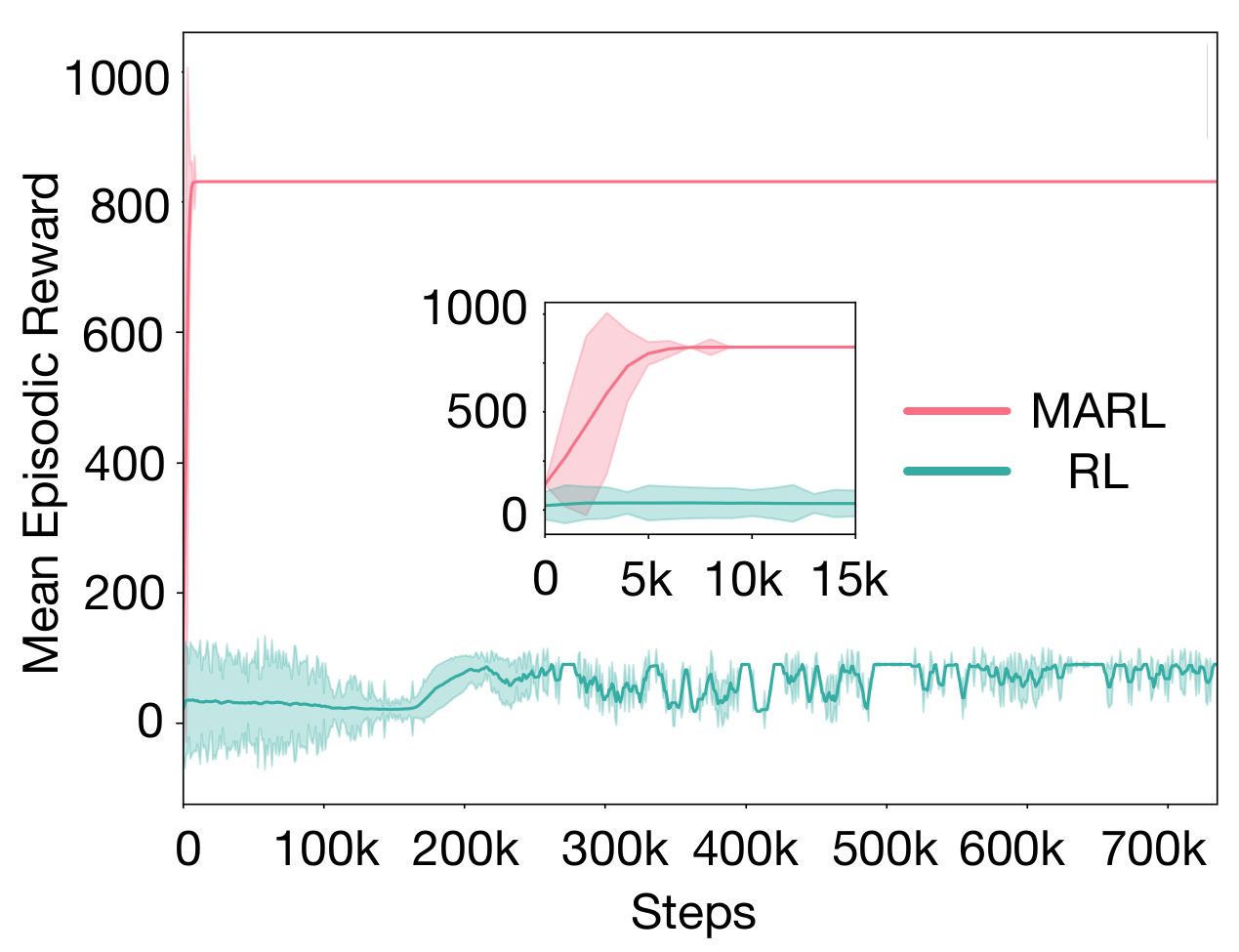}
        \caption{VGG16 Layer 2 EDP.}
        \label{fig:vgg-energy}
    \end{subfigure}

    \caption{Comparison of the MARL approach for MobileNet-v2 and VGG16 across three metrics: latency, area utilization, and EDP. MARL achieves up to 100$\times$ faster convergence than single-agent RL for MobileNet-v2 (\subref{fig:mobilenet-latency}) and 30$\times$ faster for VGG16 (\subref{fig:vgg-latency}) in minimizing latency. For area utilization, MARL shows a 40$\times$ speedup for MobileNet-v2 (\subref{fig:mobilenet-area}) and 63$\times$ for VGG16 (\subref{fig:vgg-area}). In terms of EDP, MARL outperforms single-agent RL by 300$\times$ for MobileNet-v2 (\subref{fig:mobilenet-energy}) and 41.6$\times$ for VGG16 (\subref{fig:vgg-energy}).} 

    \label{fig:2x3-comparison}
\end{figure*}

\paragraph{Actions} The action space defines mapping parameters, including categorical variables (e.g., parallelization strategies) and integers (e.g., tiling sizes). This heterogeneous space captures the complexity of DNN layer mapping. See Figure~\ref{fig:mapping_options}.

\paragraph{Observations} Each observation is a tuple $<$\texttt{latency, power, energy, area}$>$, representing a point in a multi-dimensional performance space. The agent’s objective is to navigate this space to find mappings that optimize target metrics.

\paragraph{Rewards} Rewards are designed to minimize latency, energy-delay product (EDP), and area. We use the inverse of each objective to frame the problem as maximization: $r_x = \frac{1}{X_{\text{target}}}$. For EDP, the reward uses $X_{\text{edp}} = X_{\text{latency}} \ast X_{\text{energy}}$.
By structuring the design and implementation of our environment in this way, we have developed a flexible framework for exploring the complex space of DNN mapping configurations, allowing our RL agents to learn effective strategies for optimizing critical performance metrics.


\section{Results}
\label{sec:results}
This section evaluates the effectiveness of our decentralized MARL approach for optimizing DNN mappings in hardware accelerators. We focus on three key metrics—latency, Energy-Delay Product (EDP), and area utilization—which directly impact performance, energy efficiency, and silicon cost. To assess our method, we benchmark MARL against five widely-used techniques: random search~\citep{parashar2019timeloop}, Bayesian optimization~\citep{bo-dnn-mapping}, genetic algorithms~\citep{kao2020gamma}, GAMMA~\citep{kao2020gamma}, and a single-agent RL baseline (PPO). All approaches are evaluated under equal sample budgets of 20,000, selected based on MARL's convergence profile (Section~\ref{sec:marl-rl}) to ensure fair comparison. We then assess the impact of our clustering-based agent assignment, analyze the factors contributing to MARL’s improved sample efficiency over single-agent RL, and conclude with practical guidelines for applying MARL in real-world DNN mapping tasks.


\subsection{Comparison with State-of-the-Art Search Algorithms}



\begin{figure}[t!]
  \centering
\includegraphics[width=0.87\columnwidth]{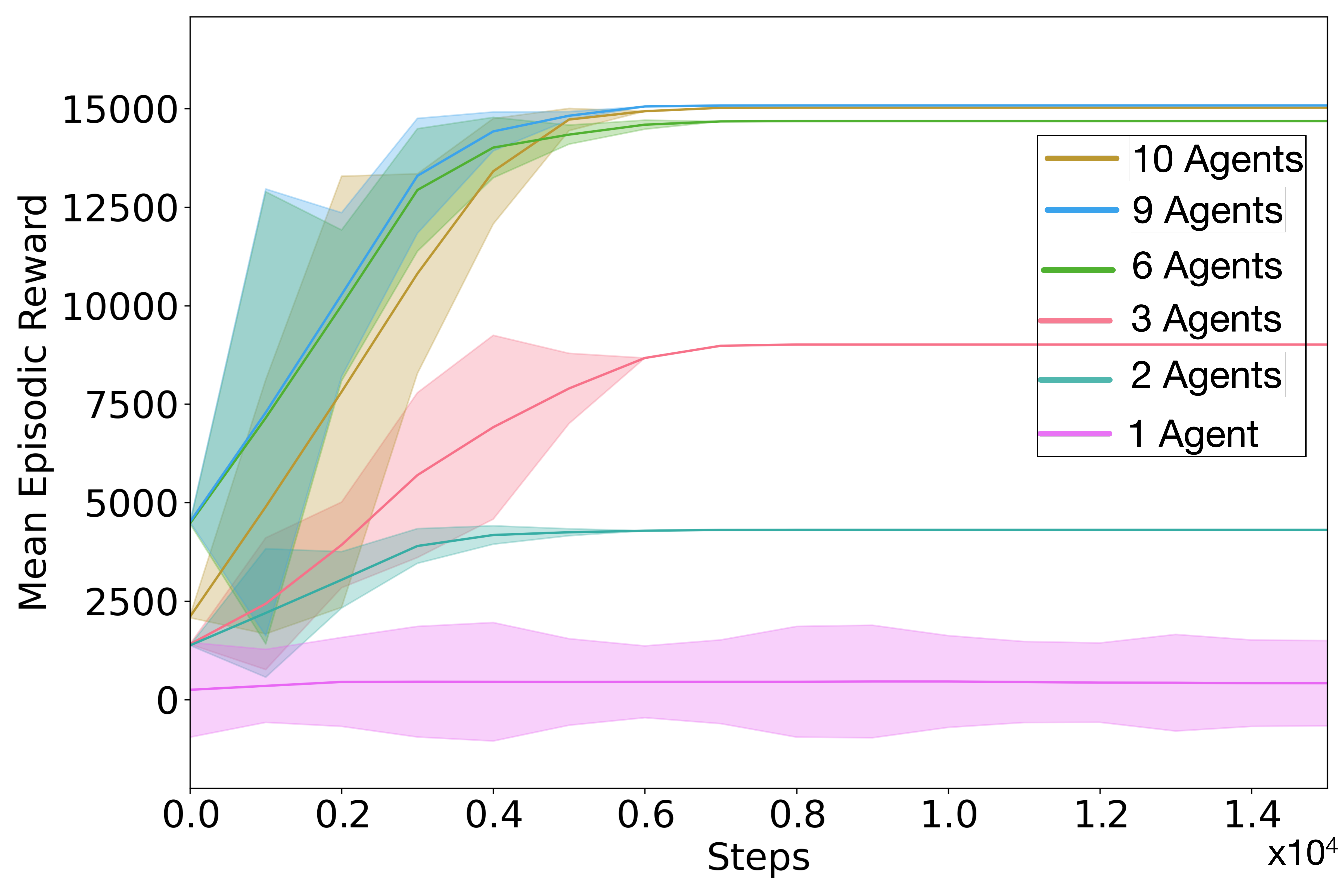}
\caption{\texttt{10 Agents} represents fully decentralized MARL with one agent per parameter. \texttt{1 Agent} is single-agent RL. \texttt{9 Agents} groups only \texttt{K} and \texttt{C} together. \texttt{2 Agents} assigns \texttt{K} and \texttt{C} to one agent and all others to another. \texttt{3 Agents} and \texttt{6 Agents} correspond to scenarios with budgets of 3 and 6 agents, respectively.\vspace{-1em}}
  \label{fig:ablation-reward}
\end{figure}%

Our evaluation encompasses two popular DNN architectures: MobileNet-v2 and VGG16 selected for their larger search space complexity. These networks were chosen for their contrasting characteristics; MobileNet-v2 represents efficient networks designed for mobile devices, while VGG16 exemplifies computationally intensive networks often deployed in high-performance computing environments. This extreme diversity allows us to assess the versatility of MARL across different network structures and computational requirements.
Table~\ref{tab:baselines} presents the comparative results for both network architectures. The data reveal that MARL consistently outperforms all baseline techniques across both network architectures. For MobileNet-v2, MARL achieves latency reductions of 3.05$\times$, 1.99$\times$, 1.33$\times$, 1.15$\times$, and 9.48$\times$ compared to random walker, Bayesian optimization, genetic algorithm, GAMMA, and single-agent RL, respectively. Similar improvements are observed for VGG16, demonstrating MARL's consistency across different network structures. In terms of energy efficiency, MARL shows even more significant gains. The improvements in EDP are particularly noteworthy, with MARL achieving up to 16.45× improvement for VGG16 compared to single-agent RL.

\subsection{Our MARL vs. Single-Agent RL Mapping Efficacy}
\label{sec:marl-rl}

Our main contribution is a decentralized MARL approach for DNN mapping. To validate its effectiveness, we compare it with single-agent RL, focusing on sample efficiency—the number of training steps required to reach optimal performance. This metric is especially relevant in DNN mapping, where large search spaces and costly simulations make efficient search critical. Unlike wall clock time, which is hardware-dependent and affected by code-level optimizations such as parallelism, sample efficiency provides a consistent, hardware-agnostic measure.

Our results show that fully decentralized MARL consistently outperforms single-agent RL in sample efficiency across all evaluated metrics.

\textbf{Latency:} Latency directly impacts quality of service in DNN applications. Figure~\ref{fig:mobilenet-latency} compares mean return, where higher rewards indicate lower latency. For MobileNet-v2, both MARL and single-agent RL converge towards similar reward levels, indicating equivalent mapping capability. However, MARL achieves this with remarkable efficiency, requiring approximately 100$\times$ fewer training steps than its single-agent counterpart. Figure~\ref{fig:vgg-latency} further highlights MARL's advantage. In this case, while single-agent RL achieves performance within 15\% of MARL's maximum attainable mean reward, it fails to fully match it. More impressively, MARL converges 30$\times$ faster than single-agent RL. This trend is consistent across other layers, including ResNet18 and AlexNet.

\textbf{Area:} Figures~\ref{fig:mobilenet-area} and~\ref{fig:vgg-area} demonstrate area-effective optimal mapping for MobileNet-v2 and VGG16 layers. The results mirror our latency findings: while both approaches converge to the same mean return (as a function of area), MARL demonstrates superior sample efficiency. For MobileNet-v2, MARL converges approximately 40$\times$ faster, while for VGG16, the speed-up is even more pronounced at 63$\times$.

\textbf{Energy Delay Product (EDP):} Figure~\ref{fig:mobilenet-energy} compares EDP optimization between MARL and single-agent RL for MobileNet. Here, MARL not only converges 312.5× faster but also achieves a 2.56× better reward (lower EDP). Similarly, for VGG16 (Figure~\ref{fig:vgg-energy}), MARL converges 41.6× faster and reaches an 8× better reward. These results demonstrate that our proposed MARL approach consistently outperforms single-agent RL in sample efficiency across various DNN layers and optimization objectives. Importantly, MARL either matches or surpasses the final performance of single-agent RL, making it a superior choice for efficient DNN mapping optimization.

\begin{figure}[t!]
  \centering
\includegraphics[width=1.0\columnwidth]{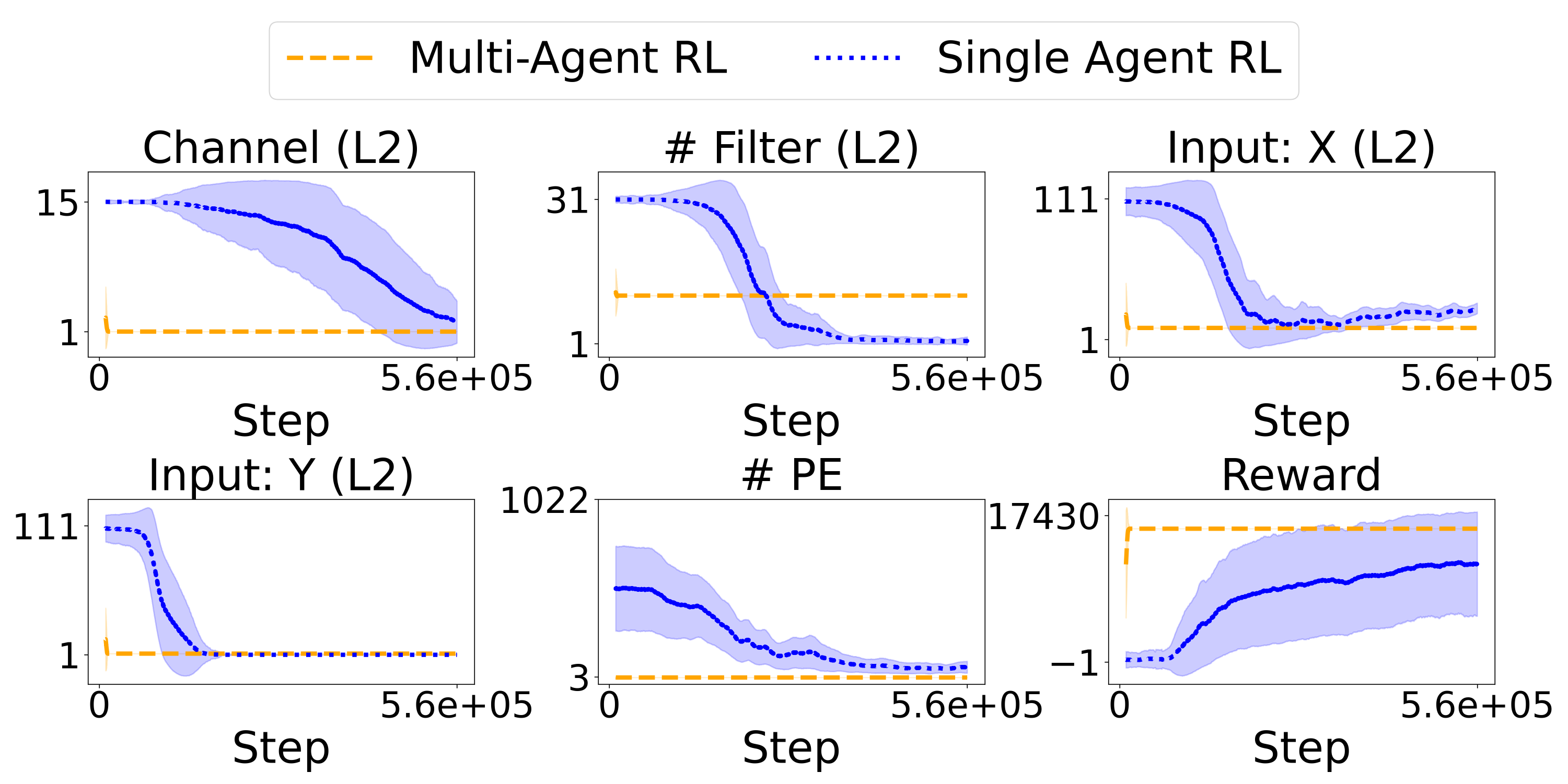}
\caption{Temporal behavior comparison of various control parameters for single agent and MARL.\vspace{-1em}}
\label{fig:action-check-mobilenet}
\end{figure}

\subsection{Adaptive Agent Assignment}

Our fully decentralized MARL approach achieves strong sample efficiency in DNN mapping (Section~\ref{sec:marl-rl}) but requires significant computational resources due to training multiple RL agents simultaneously. To address this, we integrate our clustering algorithm (Algorithm~\ref{alg:agent_assignment}) to reduce the number of agents while preserving mapping quality. The clustering algorithm identifies correlations among mapping parameters, grouping related ones under the same agent and assigning independent parameters separately. This balances control granularity with training efficiency.


To evaluate our approach, we use the MobileNet-v2 architecture as a case study, focusing on latency minimization.
Algorithm~\ref{alg:agent_assignment}, which uses correlation-based clustering to determine optimal agent assignments. 
Figure~\ref{fig:dendo-mobilenet} presents a dendrogram illustrating the relationships among various mapping parameters.
The dendrogram reveals key clustering thresholds, ranging from a single agent (\texttt{1 Agent}) to fully decentralized (\texttt{10 Agents}). These thresholds offer insights into parameter interdependencies, crucial for understanding how different aspects of the DNN architecture interact when mapped to hardware. For instance, the \texttt{9 Agents} scenario groups filters (\texttt{k}) and channels (\texttt{c}) together--- a grouping that aligns with typical hardware optimizations for convolutional layers.

Figure~\ref{fig:ablation-reward} compares the mean returns across different multi-agent scenarios, providing a quantitative basis for selecting the optimal number of agents. Our results demonstrate that MARL consistently outperforms single-agent RL (\texttt{1 Agent}). Notably, even a two-agent MARL setup (\texttt{2 Agents}) shows improved performance over the single-agent approach. We observe that increasing the number of agents leads to faster convergence, with optimal performance achieved at 9 or 6 agents. This finding has significant implications for hardware design, as it suggests that a relatively small number of specialized agents can effectively optimize complex DNN mappings.

\begin{table*}[t]
\centering
\caption{Comparison of baseline methods for MobileNet-v2 and VGG16 in terms of Latency (lower is better) and EDP (lower is better). All methods use a sample budget of 20,000 steps.}
\label{tab:baselines}
\small
\renewcommand{\arraystretch}{1.3}
\resizebox{1.7\columnwidth}{!}{%
\begin{tabular}{|c|c|c|c|c|c|c|c|c|}
\hline
\multirow{2}{*}{\textbf{Baseline Method}} 
& \multicolumn{4}{c|}{\textbf{MobileNet-v2}} 
& \multicolumn{4}{c|}{\textbf{VGG16}} \\ \cline{2-9}
& \textbf{Latency} 
& \makecell{\textbf{MARL}\\\textbf{Latency}\\\textbf{Reduction}} 
& \textbf{EDP} 
& \makecell{\textbf{MARL}\\\textbf{EDP}\\\textbf{Reduction}} 
& \textbf{Latency} 
& \makecell{\textbf{MARL}\\\textbf{Latency}\\\textbf{Reduction}} 
& \textbf{EDP} 
& \makecell{\textbf{MARL}\\\textbf{EDP}\\\textbf{Reduction}} \\ \hline\hline

Random Walker~\citep{parashar2019timeloop} 
& \cellcolor{poor}2.00E-04 & 3.05 & \cellcolor{poor}1.00E-04 & 10.00 
& \cellcolor{poor}2.94E-02 & 4.06 & \cellcolor{poor}7.87E-03 & 6.48 \\ \hline

Bayesian Optimization 
& \cellcolor{good}1.31E-04 & 1.99 & \cellcolor{good}3.33E-05 & 3.33 
& \cellcolor{good}1.08E-02 & 1.48 & \cellcolor{good}2.01E-03 & 1.65 \\ \hline

Genetic Algorithm~\citep{kao2020gamma} 
& \cellcolor{good}8.73E-05 & 1.33 & \cellcolor{good}2.50E-05 & 2.50 
& \cellcolor{good}9.26E-03 & 1.28 & \cellcolor{good}1.49E-03 & 1.22 \\ \hline

RL (PPO) 
& \cellcolor{poor}6.22E-04 & 9.48 & \cellcolor{poor}1.95E-04 & 1.95 
& \cellcolor{poor}2.36E-01 & 32.61 & \cellcolor{poor}2.00E-02 & 16.45 \\ \hline

GAMMA~\citep{kao2020gamma} 
& \cellcolor{best}7.55E-05 & 1.15 & \cellcolor{good}1.54E-05 & 0.15 
& \cellcolor{best}8.26E-03 & 1.14 & \cellcolor{good}1.41E-03 & 1.16 \\ \hline\hline

\rowcolor{thisworkrow}
\textbf{MARL (This Work)} 
& \cellcolor{best}6.56E-05 & \textbf{1.00} & \cellcolor{best}1.00E-05 & \textbf{1.00} 
& \cellcolor{best}7.25E-03 & \textbf{1.00} & \cellcolor{best}1.22E-03 & \textbf{1.00} \\ \hline
\end{tabular}%
}
\end{table*}

The efficacy of our clustering algorithm (Algorithm~\ref{alg:agent_assignment}) is evident in these results. By grouping agents based on parameter similarity, we achieve convergence rates that are competitive with, and in some cases superior to, the fully-decentralized \texttt{10 Agents} scenario. This is particularly relevant for hardware implementations, where the trade-off between optimization granularity and computational resources is a consideration.

From a hardware perspective, these findings offer valuable insights for DNN accelerator design. The ability to achieve high-quality mappings with a reduced number of agents translates to more efficient use of on-chip resources. It suggests that accelerator architectures could potentially incorporate a small number of specialized optimization units, each responsible for a group of related mapping parameters, rather than requiring a large number of independent optimization units. Moreover, the flexibility in adapting to different agent budgets aligns well with the needs of various hardware platforms, from resource-constrained edge devices to high-performance data center accelerators. 

\subsection{Mapping Decisions and Hardware Implications}
\label{sec:eval-understanding}

We hypothesize that MARL's superior sample efficiency stems from its decentralized structure, which enables parallel exploration of the design space. In this setup, each agent independently controls a specific mapping parameter, in contrast to single-agent RL, which must explore all parameters jointly. This parallelism accelerates convergence and aligns naturally with the parallel nature of DNN accelerator architectures.

To support this, we analyze the mechanisms that allow MARL to converge with fewer samples. Specifically, we perform a temporal analysis—tracking how control parameters evolve during training—to compare MARL and single-agent RL. As shown in Figure~\ref{fig:action-check-mobilenet}, both approaches eventually reach similar parameter values, but MARL does so significantly faster.

This behavior highlights MARL’s ability to converge to mapping parameters through decentralized, parallel learning—making it more sample-efficient than single-agent RL.

To further understand MARL’s sample efficiency, we analyze interdependencies among control parameters using the clustering method from Algorithm~\ref{alg:agent_assignment}. Figure~\ref{fig:correlation} shows the correlation matrix for a MobileNet layer with latency as the objective. Most parameters exhibit low correlations (0 to 0.2), indicating high independence. This supports MARL’s design of assigning separate agents to parameters for parallel optimization. From a hardware perspective, it suggests that components like memory hierarchy and PE arrangement can be tuned largely independently.


We also observe strong correlations between certain parameters—for example, \texttt{K} (filters) and \texttt{C} (channels) in MobileNet (Figure~\ref{fig:corr-mobilenet}). These insights guide our agent assignment strategy: grouping correlated parameters under the same agent and separating independent ones. This approach, as seen in the optimal 9-agent configuration, balances coordination and parallelism for more effective search.

\textit{Implication on Hardware Development.} MARL's ability to efficiently navigate complex parameter spaces has significant implications for hardware development. It suggests MARL could be a powerful tool for rapid design space exploration, speeding up DNN accelerator development by quickly identifying promising regions. This would enable hardware designers to focus efforts more efficiently, leading to faster iterations and more optimized designs.

\subsection{Implementation and Overhead Considerations}


While our MARL approach significantly improves sample efficiency, it incurs an initial overhead for dataset curation of about 20,000 samples to estimate correlations and assign agents. This overhead was excluded from the 30-300$\times$ efficiency comparisons in Section~\ref{sec:marl-rl}. However, considering the overall gains, this overhead remains small. Single-agent RL typically requires hundreds of thousands of steps to converge, while MARL, excluding the overhead, converges in around 5,000 samples. Even including the 20,000-sample overhead, MARL’s total sample requirement of approximately 25,000 is still an order of magnitude lower than single-agent RL.


From a hardware design perspective, this upfront cost is a fixed investment that can be amortized over multiple optimization runs or accelerator designs. For instance, optimizing an accelerator for various DNN models or through iterative design cycles quickly renders the initial cost negligible. Additionally, the dataset curation provides insights into parameter interdependencies, informing architectural decisions and offering deeper understanding of which hardware resources might benefit from closer coupling or increased flexibility.




In practice, our MARL approach integrates seamlessly into modern hardware design workflows. The process begins with design space exploration, using a 20,000-sample dataset for correlation analysis and agent assignment to establish a baseline understanding. MARL’s fast convergence—typically within 5,000 samples—enables rapid iteration and efficient exploration of design points. Insights from multiple runs can then be used to refine the accelerator architecture, aligning it more closely with learned parameter relationships. Finally, the learned correlations and agent assignments can be generalized to similar DNN architectures, further amortizing the initial data collection cost. While dataset curation introduces some overhead, its impact is minimal compared to the overall efficiency gains achieved in typical DNN mapping scenarios.




\section{Conclusion}
This paper presents the first known implementation of a decentralized Multi-Agent Reinforcement Learning (MARL) approach to address the complex problem of efficiently mapping deep neural network layers onto specific hardware platforms to optimize latency, energy consumption, and area utilization. We contribute an algorithm for agent assignment based on correlation and clustering. Our work demonstrates a strong improvement in sample efficiency, achieving 30-300$\times$ faster convergence compared to single-agent RL and state of the art. 

\sloppy
\bibliographystyle{ACM-Reference-Format}

\bibliography{sample-base}


\begin{thebibliography}{25}


\ifx \showCODEN    \undefined \def \showCODEN     #1{\unskip}     \fi
\ifx \showDOI      \undefined \def \showDOI       #1{#1}\fi
\ifx \showISBNx    \undefined \def \showISBNx     #1{\unskip}     \fi
\ifx \showISBNxiii \undefined \def \showISBNxiii  #1{\unskip}     \fi
\ifx \showISSN     \undefined \def \showISSN      #1{\unskip}     \fi
\ifx \showLCCN     \undefined \def \showLCCN      #1{\unskip}     \fi
\ifx \shownote     \undefined \def \shownote      #1{#1}          \fi
\ifx \showarticletitle \undefined \def \showarticletitle #1{#1}   \fi
\ifx \showURL      \undefined \def \showURL       {\relax}        \fi
\providecommand\bibfield[2]{#2}
\providecommand\bibinfo[2]{#2}
\providecommand\natexlab[1]{#1}
\providecommand\showeprint[2][]{arXiv:#2}

\bibitem[Brockman et~al\mbox{.}(2016)]%
        {gym}
\bibfield{author}{\bibinfo{person}{Greg Brockman}, \bibinfo{person}{Vicki Cheung}, \bibinfo{person}{Ludwig Pettersson}, \bibinfo{person}{Jonas Schneider}, \bibinfo{person}{John Schulman}, \bibinfo{person}{Jie Tang}, {and} \bibinfo{person}{Wojciech Zaremba}.} \bibinfo{year}{2016}\natexlab{}.
\newblock \showarticletitle{OpenAI Gym}.
\newblock \bibinfo{journal}{\emph{CoRR}}  \bibinfo{volume}{abs/1606.01540} (\bibinfo{year}{2016}).
\newblock
\showeprint[arXiv]{1606.01540}
\urldef\tempurl%
\url{http://arxiv.org/abs/1606.01540}
\showURL{%
\tempurl}


\bibitem[Chai et~al\mbox{.}(2021)]%
        {deeplearning-cv}
\bibfield{author}{\bibinfo{person}{Junyi Chai}, \bibinfo{person}{Hao Zeng}, \bibinfo{person}{Anming Li}, {and} \bibinfo{person}{Eric~WT Ngai}.} \bibinfo{year}{2021}\natexlab{}.
\newblock \showarticletitle{Deep learning in computer vision: A critical review of emerging techniques and application scenarios}.
\newblock \bibinfo{journal}{\emph{Machine Learning with Applications}}  \bibinfo{volume}{6} (\bibinfo{year}{2021}), \bibinfo{pages}{100134}.
\newblock


\bibitem[Chen et~al\mbox{.}(2016)]%
        {dnn-mapping2}
\bibfield{author}{\bibinfo{person}{Yu-Hsin Chen}, \bibinfo{person}{Joel Emer}, {and} \bibinfo{person}{Vivienne Sze}.} \bibinfo{year}{2016}\natexlab{}.
\newblock \showarticletitle{Eyeriss: A spatial architecture for energy-efficient dataflow for convolutional neural networks}.
\newblock \bibinfo{journal}{\emph{ACM SIGARCH computer architecture news}} \bibinfo{volume}{44}, \bibinfo{number}{3} (\bibinfo{year}{2016}), \bibinfo{pages}{367--379}.
\newblock


\bibitem[Choquette et~al\mbox{.}(2021)]%
        {choquette2021nvidia}
\bibfield{author}{\bibinfo{person}{Jack Choquette}, \bibinfo{person}{Wishwesh Gandhi}, \bibinfo{person}{Olivier Giroux}, \bibinfo{person}{Nick Stam}, {and} \bibinfo{person}{Ronny Krashinsky}.} \bibinfo{year}{2021}\natexlab{}.
\newblock \showarticletitle{NVIDIA A100 tensor core GPU: Performance and innovation}.
\newblock \bibinfo{journal}{\emph{IEEE Micro}} \bibinfo{volume}{41}, \bibinfo{number}{2} (\bibinfo{year}{2021}), \bibinfo{pages}{29--35}.
\newblock


\bibitem[Clustering(2023)]%
        {agglo-clustering-2}
\bibfield{author}{\bibinfo{person}{Clustering}.} \bibinfo{year}{2023}\natexlab{}.
\newblock \bibinfo{booktitle}{\emph{Log-linear Models for Contingency Tables}}.
\newblock
\urldef\tempurl%
\url{https://online.stat.psu.edu/stat555/node/86/}
\showURL{%
\tempurl}
\newblock
\shownote{Accessed: 2023-10-19}.


\bibitem[et~al.(2017a)]%
        {mobilenet}
\bibfield{author}{\bibinfo{person}{Howard et al.}} \bibinfo{year}{2017}\natexlab{a}.
\newblock \showarticletitle{MobileNets: Efficient Convolutional Neural Networks for Mobile Vision Applications}.
\newblock \bibinfo{journal}{\emph{CoRR}}  \bibinfo{volume}{abs/1704.04861} (\bibinfo{year}{2017}).
\newblock
\showeprint[arXiv]{1704.04861}
\urldef\tempurl%
\url{http://arxiv.org/abs/1704.04861}
\showURL{%
\tempurl}


\bibitem[et~al.(2017b)]%
        {tpu}
\bibfield{author}{\bibinfo{person}{Jouppi et al.}} \bibinfo{year}{2017}\natexlab{b}.
\newblock \showarticletitle{In-datacenter performance analysis of a tensor processing unit}. In \bibinfo{booktitle}{\emph{Proceedings of the 44th annual international symposium on computer architecture}}. \bibinfo{pages}{1--12}.
\newblock


\bibitem[et~al.(2021)]%
        {mindmappings}
\bibfield{author}{\bibinfo{person}{Kartik et al.}} \bibinfo{year}{2021}\natexlab{}.
\newblock \showarticletitle{Mind Mappings: Enabling Efficient Algorithm-Accelerator Mapping Space Search}. In \bibinfo{booktitle}{\emph{Proceedings of the 26th ACM International Conference on Architectural Support for Programming Languages and Operating Systems}} (Virtual, USA) \emph{(\bibinfo{series}{ASPLOS '21})}. \bibinfo{publisher}{Association for Computing Machinery}, \bibinfo{address}{New York, NY, USA}, \bibinfo{pages}{943–958}.
\newblock
\showISBNx{9781450383172}
\urldef\tempurl%
\url{https://doi.org/10.1145/3445814.3446762}
\showDOI{\tempurl}


\bibitem[et~al.(2022)]%
        {llm1}
\bibfield{author}{\bibinfo{person}{Long et al.}} \bibinfo{year}{2022}\natexlab{}.
\newblock \bibinfo{title}{Training language models to follow instructions with human feedback}.
\newblock
\newblock
\showeprint[arxiv]{2203.02155}~[cs.CL]


\bibitem[et~al.(2019a)]%
        {parashar2019timeloop}
\bibfield{author}{\bibinfo{person}{Parashar et al.}} \bibinfo{year}{2019}\natexlab{a}.
\newblock \showarticletitle{Timeloop: A systematic approach to dnn accelerator evaluation}. In \bibinfo{booktitle}{\emph{2019 IEEE international symposium on performance analysis of systems and software (ISPASS)}}. IEEE, \bibinfo{pages}{304--315}.
\newblock


\bibitem[et~al.(2019b)]%
        {megatron}
\bibfield{author}{\bibinfo{person}{Shoeybi et al.}} \bibinfo{year}{2019}\natexlab{b}.
\newblock \showarticletitle{Megatron-LM: Training Multi-Billion Parameter Language Models Using Model Parallelism}.
\newblock \bibinfo{journal}{\emph{CoRR}}  \bibinfo{volume}{abs/1909.08053} (\bibinfo{year}{2019}).
\newblock
\showeprint[arXiv]{1909.08053}
\urldef\tempurl%
\url{http://arxiv.org/abs/1909.08053}
\showURL{%
\tempurl}


\bibitem[He et~al\mbox{.}(2015)]%
        {resnet-cv}
\bibfield{author}{\bibinfo{person}{Kaiming He}, \bibinfo{person}{Xiangyu Zhang}, \bibinfo{person}{Shaoqing Ren}, {and} \bibinfo{person}{Jian Sun}.} \bibinfo{year}{2015}\natexlab{}.
\newblock \showarticletitle{Deep Residual Learning for Image Recognition}.
\newblock \bibinfo{journal}{\emph{CoRR}}  \bibinfo{volume}{abs/1512.03385} (\bibinfo{year}{2015}).
\newblock
\showeprint[arXiv]{1512.03385}
\urldef\tempurl%
\url{http://arxiv.org/abs/1512.03385}
\showURL{%
\tempurl}


\bibitem[Kao and Krishna(2020)]%
        {kao2020gamma}
\bibfield{author}{\bibinfo{person}{Sheng-Chun Kao} {and} \bibinfo{person}{Tushar Krishna}.} \bibinfo{year}{2020}\natexlab{}.
\newblock \showarticletitle{Gamma: Automating the hw mapping of dnn models on accelerators via genetic algorithm}. In \bibinfo{booktitle}{\emph{Proceedings of the 39th International Conference on Computer-Aided Design}}. \bibinfo{pages}{1--9}.
\newblock


\bibitem[Krishnan et~al\mbox{.}(2023)]%
        {archgym}
\bibfield{author}{\bibinfo{person}{Srivatsan Krishnan}, \bibinfo{person}{Amir Yazdanbakhsh}, \bibinfo{person}{Shvetank Prakash}, \bibinfo{person}{Jason Jabbour}, \bibinfo{person}{Ikechukwu Uchendu}, \bibinfo{person}{Susobhan Ghosh}, \bibinfo{person}{Behzad Boroujerdian}, \bibinfo{person}{Daniel Richins}, \bibinfo{person}{Devashree Tripathy}, \bibinfo{person}{Aleksandra Faust}, {and} \bibinfo{person}{Vijay Janapa~Reddi}.} \bibinfo{year}{2023}\natexlab{}.
\newblock \showarticletitle{ArchGym: An Open-Source Gymnasium for Machine Learning Assisted Architecture Design}. In \bibinfo{booktitle}{\emph{Proceedings of the 50th Annual International Symposium on Computer Architecture}} (Orlando, FL, USA) \emph{(\bibinfo{series}{ISCA '23})}. \bibinfo{publisher}{Association for Computing Machinery}, \bibinfo{address}{New York, NY, USA}, Article \bibinfo{articleno}{14}, \bibinfo{numpages}{16}~pages.
\newblock
\showISBNx{9798400700958}
\urldef\tempurl%
\url{https://doi.org/10.1145/3579371.3589049}
\showDOI{\tempurl}


\bibitem[Krizhevsky et~al\mbox{.}(2012)]%
        {alexnet}
\bibfield{author}{\bibinfo{person}{Alex Krizhevsky}, \bibinfo{person}{Ilya Sutskever}, {and} \bibinfo{person}{Geoffrey~E Hinton}.} \bibinfo{year}{2012}\natexlab{}.
\newblock \showarticletitle{ImageNet Classification with Deep Convolutional Neural Networks}.
\newblock   \bibinfo{volume}{25} (\bibinfo{year}{2012}).
\newblock


\bibitem[Kwon et~al\mbox{.}(2020)]%
        {dnn-mapping1}
\bibfield{author}{\bibinfo{person}{Hyoukjun Kwon}, \bibinfo{person}{Prasanth Chatarasi}, \bibinfo{person}{Vivek Sarkar}, \bibinfo{person}{Tushar Krishna}, \bibinfo{person}{Michael Pellauer}, {and} \bibinfo{person}{Angshuman Parashar}.} \bibinfo{year}{2020}\natexlab{}.
\newblock \showarticletitle{MAESTRO: A Data-Centric Approach to Understand Reuse, Performance, and Hardware Cost of DNN Mappings}.
\newblock \bibinfo{journal}{\emph{IEEE Micro}} \bibinfo{volume}{40}, \bibinfo{number}{3} (\bibinfo{year}{2020}), \bibinfo{pages}{20--29}.
\newblock
\urldef\tempurl%
\url{https://doi.org/10.1109/MM.2020.2985963}
\showDOI{\tempurl}


\bibitem[Lukasov{\'a}(1979)]%
        {agglo-clustering}
\bibfield{author}{\bibinfo{person}{Alena Lukasov{\'a}}.} \bibinfo{year}{1979}\natexlab{}.
\newblock \showarticletitle{Hierarchical agglomerative clustering procedure}.
\newblock \bibinfo{journal}{\emph{Pattern Recognition}} \bibinfo{volume}{11}, \bibinfo{number}{5-6} (\bibinfo{year}{1979}), \bibinfo{pages}{365--381}.
\newblock


\bibitem[Parekkadan and Das(2022)]%
        {dnn-mapping-rl}
\bibfield{author}{\bibinfo{person}{S Parekkadan} {and} \bibinfo{person}{S Das}.} \bibinfo{year}{2022}\natexlab{}.
\newblock \showarticletitle{Reinforcement Learning-based Efficient Mapping of DNN Models onto Accelerators}. In \bibinfo{booktitle}{\emph{Proceedings of the IEEE Symposium on Low-Power and High-Speed Chips (COOL CHIPS)}}. \bibinfo{pages}{1--6}.
\newblock


\bibitem[Sandler et~al\mbox{.}(2018)]%
        {mobilenet-v2}
\bibfield{author}{\bibinfo{person}{Mark Sandler}, \bibinfo{person}{Andrew~G. Howard}, \bibinfo{person}{Menglong Zhu}, \bibinfo{person}{Andrey Zhmoginov}, {and} \bibinfo{person}{Liang{-}Chieh Chen}.} \bibinfo{year}{2018}\natexlab{}.
\newblock \showarticletitle{Inverted Residuals and Linear Bottlenecks: Mobile Networks for Classification, Detection and Segmentation}.
\newblock \bibinfo{journal}{\emph{CoRR}}  \bibinfo{volume}{abs/1801.04381} (\bibinfo{year}{2018}).
\newblock
\showeprint[arXiv]{1801.04381}
\urldef\tempurl%
\url{http://arxiv.org/abs/1801.04381}
\showURL{%
\tempurl}


\bibitem[Shen et~al\mbox{.}(2017)]%
        {grid-search-dnn1}
\bibfield{author}{\bibinfo{person}{Yongming Shen}, \bibinfo{person}{Michael Ferdman}, {and} \bibinfo{person}{Peter Milder}.} \bibinfo{year}{2017}\natexlab{}.
\newblock \showarticletitle{Maximizing CNN Accelerator Efficiency Through Resource Partitioning}.
\newblock \bibinfo{journal}{\emph{SIGARCH Comput. Archit. News}} \bibinfo{volume}{45}, \bibinfo{number}{2} (\bibinfo{date}{jun} \bibinfo{year}{2017}), \bibinfo{pages}{535–547}.
\newblock
\showISSN{0163-5964}
\urldef\tempurl%
\url{https://doi.org/10.1145/3140659.3080221}
\showDOI{\tempurl}


\bibitem[Shi et~al\mbox{.}(2020)]%
        {bo-dnn-mapping}
\bibfield{author}{\bibinfo{person}{Zhan Shi}, \bibinfo{person}{Chirag Sakhuja}, \bibinfo{person}{Milad Hashemi}, \bibinfo{person}{Kevin Swersky}, {and} \bibinfo{person}{Calvin Lin}.} \bibinfo{year}{2020}\natexlab{}.
\newblock \showarticletitle{Using bayesian optimization for hardware/software co-design of neural accelerators}. In \bibinfo{booktitle}{\emph{Workshop on ML for Systems at the Conference on Neural Information Processing Systems (NeurIPS)}}.
\newblock


\bibitem[Simonyan and Zisserman(2015)]%
        {vgg}
\bibfield{author}{\bibinfo{person}{Karen Simonyan} {and} \bibinfo{person}{Andrew Zisserman}.} \bibinfo{year}{2015}\natexlab{}.
\newblock \bibinfo{title}{Very Deep Convolutional Networks for Large-Scale Image Recognition}.
\newblock
\newblock
\showeprint[arxiv]{1409.1556}~[cs.CV]


\bibitem[Stoutchinin et~al\mbox{.}(2019)]%
        {grid-serach-dnn2}
\bibfield{author}{\bibinfo{person}{Arthur Stoutchinin}, \bibinfo{person}{Francesco Conti}, {and} \bibinfo{person}{Luca Benini}.} \bibinfo{year}{2019}\natexlab{}.
\newblock \showarticletitle{Optimally scheduling CNN convolutions for efficient memory access}.
\newblock \bibinfo{journal}{\emph{arXiv preprint arXiv:1902.01492}} (\bibinfo{year}{2019}).
\newblock


\bibitem[Suda et~al\mbox{.}(2016)]%
        {grid-search-dnn3}
\bibfield{author}{\bibinfo{person}{Naveen Suda}, \bibinfo{person}{Vikas Chandra}, \bibinfo{person}{Ganesh Dasika}, \bibinfo{person}{Abinash Mohanty}, \bibinfo{person}{Yufei Ma}, \bibinfo{person}{Sarma Vrudhula}, \bibinfo{person}{Jae-sun Seo}, {and} \bibinfo{person}{Yu Cao}.} \bibinfo{year}{2016}\natexlab{}.
\newblock \showarticletitle{Throughput-optimized OpenCL-based FPGA accelerator for large-scale convolutional neural networks}. In \bibinfo{booktitle}{\emph{Proceedings of the 2016 ACM/SIGDA international symposium on field-programmable gate arrays}}. \bibinfo{pages}{16--25}.
\newblock


\bibitem[Sutton and Barto(2018)]%
        {sutton2018reinforcement}
\bibfield{author}{\bibinfo{person}{Richard~S Sutton} {and} \bibinfo{person}{Andrew~G Barto}.} \bibinfo{year}{2018}\natexlab{}.
\newblock \bibinfo{booktitle}{\emph{Reinforcement learning: An introduction}}.
\newblock


\end{thebibliography}

\vspace{12pt}
\color{red}

\end{document}